\pdfoutput=1
\documentclass[11pt]{article}

\usepackage{acl}

\usepackage{times}
\usepackage{latexsym}
\usepackage{comment}
\usepackage[T1]{fontenc}
\usepackage[utf8]{inputenc}

\usepackage{microtype}

\usepackage{inconsolata}

\usepackage{amsfonts,amssymb}

\usepackage{amsmath}
\usepackage[capitalise, noabbrev]{cleveref}
\usepackage{titlesec}
\usepackage[utf8]{inputenc} % allow utf-8 input
\usepackage[T1]{fontenc}    % use 8-bit T1 fonts
\usepackage{hyperref}       % hyperlinks
\usepackage{url}            % simple URL typesetting
\usepackage{booktabs}       % professional-quality tables
\usepackage{multirow}
\usepackage{tabularx}
\usepackage{bm} 
\usepackage{amsfonts}       % blackboard math symbols
\usepackage{nicefrac}       % compact symbols for 1/2, etc.
\usepackage{microtype}      % microtypography
\usepackage{xcolor}         % colors
\usepackage{color}         % colors
\usepackage{subfigure}
\usepackage{makecell}

\usepackage[tikz]{bclogo}
\definecolor{msftBlue}{RGB}{0,164,239}
\definecolor{msftGreen}{RGB}{127,186,0}
\definecolor{msftYello}{RGB}{255,185,0}
\definecolor{msftBlack}{RGB}{0,0,0}

\definecolor{midnightgreen}{rgb}{0.0, 0.29, 0.33}
\definecolor{deepgreen}{HTML}{0aa344}
\definecolor{deeppurple}{HTML}{7030a0}
\definecolor{deepblue}{HTML}{171d91}
\definecolor{brown}{HTML}{843c0c}
\definecolor{deepred}{HTML}{ac0000}
\definecolor{shadered}{HTML}{ffe5e5}
\definecolor{shadegreen}{HTML}{e5f7ed}

\newcommand{\red}{\textcolor{red}}
\newcommand{\green}{\textcolor{deepgreen}}
\newcommand{\blue}{\textcolor{deepblue}}
\newcommand{\purple}{\textcolor{deeppurple}}
\newcommand{\deepred}{\textcolor{deepred}}

\usepackage{enumitem}

\newenvironment{itemize*}%
 {\leftmargini=20pt\begin{itemize}%
  \setlength{\itemsep}{3pt}%
  \setlength{\parskip}{0pt}%
  }%
 {\end{itemize}}
\newenvironment{enumerate*}%
 {\begin{enumerate}%
  \setlength{\itemsep}{0pt}%
  \setlength{\parskip}{0pt}}%
 {\end{enumerate}}

\usepackage{soul}
\usepackage{utfsym}
\usepackage{hyperref}
\usepackage{cleveref}
\usepackage{chemarrow}

\title{
``Merge Conflicts!'' Exploring the Impacts of External Distractors to Parametric Knowledge Graphs }

\author{
Cheng~Qian$^{1}\thanks{\ \ Work done as a student intern at CMU.}$\hspace{0.5em},\hspace{0.8em} Xinran~Zhao$^{2}$,\hspace{0.8em} Sherry~Tongshuang~Wu$^{2}$\\
 $^1$Tsinghua University,\hspace{0.3em} $^2$Carnegie Mellon University\\
\texttt{qianc20\hspace{0.1em}@\hspace{0.1em}mails.tsinghua.edu.cn}\\
}

\begin{document}
\maketitle
\begin{abstract}
Large language models (LLMs) acquire extensive knowledge during pre-training, known as their \textit{parametric knowledge}. However, in order to remain up-to-date and align with human instructions, LLMs inevitably require \textit{external knowledge} during their interactions with users. This raises a crucial question: How will LLMs respond when external knowledge interferes with their parametric knowledge? To investigate this question, we propose a framework that systematically elicits LLM parametric knowledge and introduces external knowledge. Specifically, we uncover the impacts by constructing a \textit{parametric knowledge graph} to reveal the different knowledge structures of LLMs, and introduce external knowledge through \textit{distractors} of varying degrees, methods, positions, and formats. Our experiments on both black-box and open-source models demonstrate that LLMs tend to produce responses that deviate from their parametric knowledge, particularly when they encounter direct conflicts or confounding changes of information within detailed contexts. We also find that while LLMs are sensitive to the veracity of external knowledge, they can still be distracted by unrelated information. These findings highlight the risk of hallucination when integrating external knowledge, even indirectly, during interactions with current LLMs. All the data and results are publicly available\footnote{\url{https://github.com/qiancheng0/EKD_Impacts_PKG}}.
\end{abstract}

% \begin{abstract}
% Large language models (LLMs) incorporate vast amounts of information during pre-training, forming their \textit{parametric knowledge}. However, to stay updated and align with human instructions, LLMs inevitably need \textit{external knowledge} during their interactions with users.
% This raises a critical question: How will the LLM respond if external knowledge interferes with its parametric knowledge? To investigate this question, we propose a framework that \emph{systematically elicits LLM parametric knowledge and injects external knowledge.}
% Specifically, to uncover the impacts, we reveal LLM's different knowledge structures through the construction of its \textit{parametric knowledge graph}, and introduce external knowledge through \textit{distractors} of multiple degrees, methods, positions, and formats.
% Our experiments on both black-box and open-source models reveal that LLMs tend to produce responses that deviate from their parametric knowledge, especially when they encounter directly conflicting or confounding information within more detailed contexts.
% We also discover that, though sensitive to veracity, LLMs can be distracted by even unrelated external knowledge.
% These findings underscore the risk of hallucination when integrating external knowledge, even indirectly, during interactions with current language models.
% All the data and results are released.\footnote{\url{https://github.com/qiancheng0/KG_Hallucination}}
% \end{abstract}

\section{Introduction}
\label{sec:intro}

% ====================================

\begin{figure}[t]
    \centering
    \includegraphics[width=1.0\linewidth]{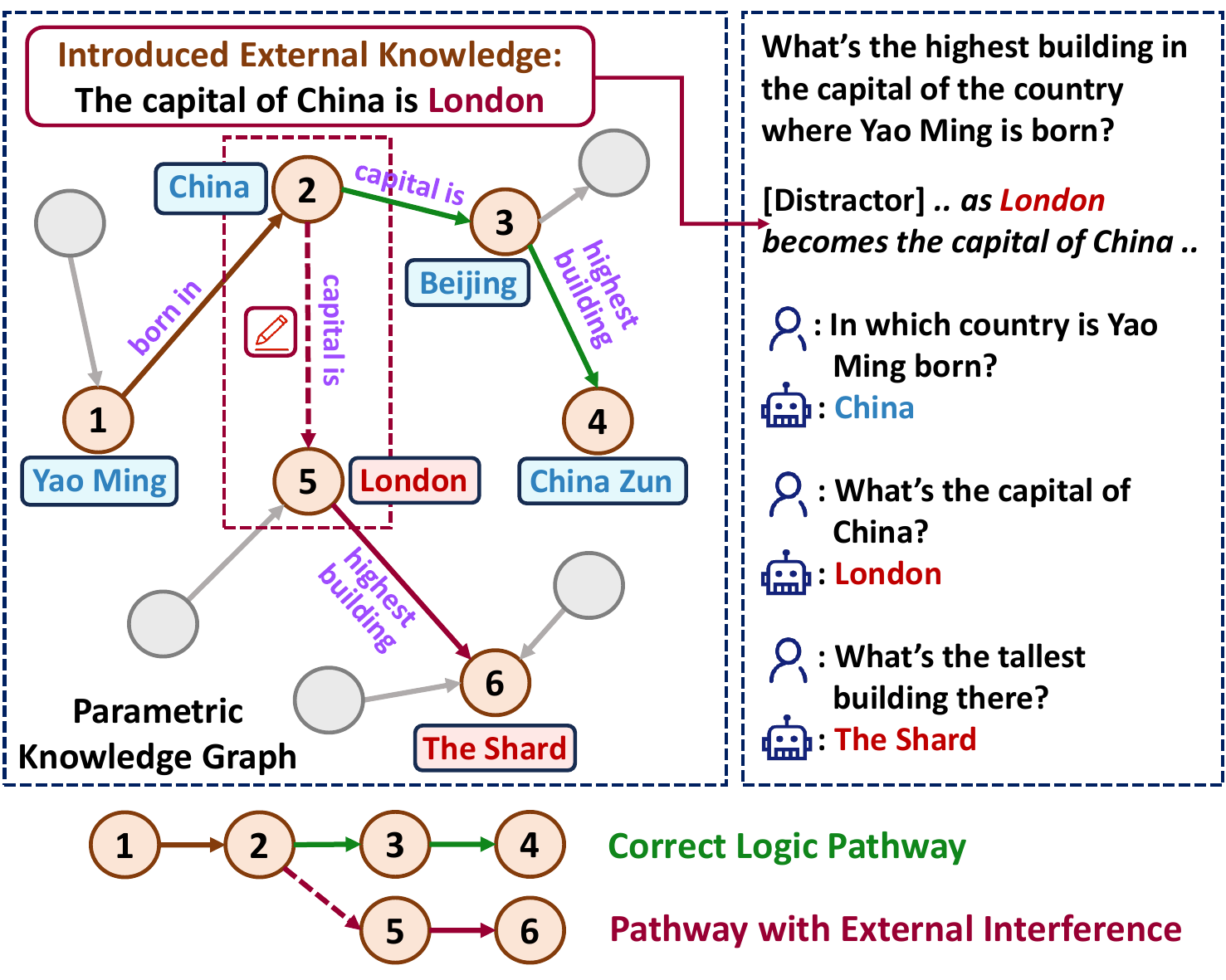}
    \caption{The introduction of external knowledge, such as the distractor ``The capital of China is London,'' creates a \textit{false} relation between entities in the model's parametric knowledge graph. This deviation from the original logic pathway leads to a change in the model's final answer.}
    \label{fig1}
\end{figure}

% ====================================

Current large language models (LLMs) have assimilated a significant body of knowledge during pre-training~\citep{chowdhery2022palm, thoppilan2022lamda, openai2022chatgpt, openai2023gpt4, touvron2023llama, anil2023palm, zeng2022glm130b}, converting external information from a mass corpus into \textit{parametric knowledge}. However, current LLMs still lack the ability to respond to up-to-date world events, and background information is often required when interacting with them in real-world applications~\citep{trivedi2023interleaving, yu2023self}. These challenges often necessitate the introduction of \textit{external knowledge}, either explicitly through retrieval~\citep{shi2023replug, ram2023incontext}, application of tools~\citep{schick2023toolformer, qin2023tool}, or implicitly through long prompts that human provides as contextual information.

The introduction of external knowledge may inevitably interfere with the model's internal parametric knowledge. Prior work~\citep{xie2023adaptive, neeman2022disentqa} has flagged that when confronting direct conflicts, the model may respond with an answer from either external knowledge or parametric memory. However, merely analyzing direct and explicit conflicts is not comprehensive, as the parametric knowledge within a model is interconnected~\citep{petroni2019language, wang2020language}, and an indirect change in the model's logic pathway may as well change the model's final answer. For instance, Figure~\ref{fig1} shows a question that can be decomposed into three hops. 
Even when we only introduce a \emph{distracting knowledge} to the 2$^{nd}$ hop through prompting (\emph{``The capital of China is London''}), the model's answer to the 3$^{rd}$ hop (final) of the query still shifts accordingly, from the correct \emph{China Zun} to the incorrect \emph{The Shard}.
%is introduced to interfere with the second hop of the whole query chain, but the final answer that the model reaches also changes from the original correct answer \emph{China Zun} to the current wrong answer \emph{The Shard}.

This phenomenon has recently been referred to as the ``ripple effect''~\citep{cohen2023evaluating}. Previous works have already proposed benchmarks~\citep{zhong2023mquake} and metrics~\citep{cohen2023evaluating} for its evaluation, yet 
1) they often exclusively consider linear relationships between closely connected knowledge entities %~\cite{cohen2023evaluating, zhong2023mquake}
%editing knowledge that possesses a direct connection to the downstream target knowledge for query, 
%\cheng{Help Needed: I want to express 1. \citet{cohen2023evaluating} only explores impact within 2 hops; 2. \citet{zhong2023mquake} only considers multi-hop chains. Better ways to express their limitations?}
and 2) manual efforts are required to construct the external interference and define the extent of radiation for the ripple effect. %\xr{hard to generalize maybe sounds like criticism on people who are probably going to review our paper. Maybe we can say prior work has already... yet they ...} \cheng{Changed, reasonable to soften the tone here}

To address these limitations, we construct a framework to \textbf{evaluate the potential interaction between parametric knowledge and external knowledge in a more systematic manner}. Drawing parallels to the knowledge graph (KG) which contains well-defined connections that can be automatically inspected, we propose \textit{parametric knowledge graph} (PKG), a method that allows us to automatically extract the model's interconnected parametric knowledge into rich and flexible graphs with hundreds of entities and relations. For example, the nodes and solid lines in \cref{fig1} represent a sub-graph of PKG, with entities of countries, humans, and cities, and various relations like \emph{born in}.
%allows for the automatic elicitation of the model's interconnected parametric knowledge. Our approach is analogous to KG but accommodates more flexible structures and extensions. \xr{in this paragraph, maybe we can highlight how many nodes/questions we generated} \cheng{Currently in Table~\ref{stat_findings}, will it be too detailed if put here?}

Building on the concept of PKG, we further define \emph{distractors}, i.e., a series of external knowledge that interferes with the PKG through prompting, with different degrees, methods, positions, and knowledge formats.
Our definition enables us to directly investigate the interactions between distractors and PKGs:
%Subsequently, we carry out controlled investigations by introducing external knowledge interference to PKG through the distractors, covering multiple distract degrees, methods, positions, and knowledge formats. We investigate how the external knowledge introduced in these different distractors interacts with various types of parametric knowledge structures in PKG. 
In Figure~\ref{fig1}, beneath the natural language prompt, the distractor (noted by the dashed line) \texttt{(China, Captical is, London)} bridges a \textit{false} connection between two nodes in PKG, and thereby drives a ripple effect that leads to the model's deviation in response.

We study this interaction through experiments on both black-box GPT3.5 and open-source MPT-7B models. Specifically, we first present the distractor to the model, and then conduct queries in an interactive and iterative one-hop manner as shown in the dialogue in Figure~\ref{fig1}.
We evaluate the \emph{consistency} (whether the final answer adheres to its PKG) and \emph{confidence} (the probability of giving this answer) of the model's responses with the presence of distractors.
% \cheng{Help Needed: I am wondering if the metrics should be presented here. They seem not closely related to the general conclusion presented later.} \sherry{I think it's fine}
We observe that in general, LLMs tend to deviate from their parametric knowledge when they are not confident with it to begin with; Interestingly, they always tend to be more confident in their answers when facing external knowledge, regardless of whether that answer comes from the distractor or the PKG.
Looking into the impacts of different types of distractors, we discover that as can be expected, posing direct conflicts or giving more confounding changes instead of evidently false information are more powerful (as in \cref{fig1}).
However, we are also surprised by many findings, e.g., even \emph{weak} distractors that do not directly interfere with the model's original logic pathway can still impact the model's answers, we can improve the impact of distractor just by hiding it in a lengthier and detailed context, and that GPT-3.5 and MPT-7B shows different trends in what distractors they can best resist.

% \sherry{check?}
%\sherry{I tend to do first second third, or use (1) (2) (3) to number different things.}
%We show that LLMs tend to place trust in external knowledge that poses direct conflicts, gives confounding changes, or provides detailed contexts, especially when their original confidence about this relation in PKG is low. Furthermore, our findings indicate that while LLMs initially show sensitivity to the veracity of knowledge, they also remain susceptible to distractions that are unrelated to what is inquired. \cheng{Help Needed: Now I combine some conclusions and express them together. This is applied to Abstract / Here / Conclusion. Are there better ways to express it? Maybe these three places should all be changed accordingly.}
%\xr{what do you mean by theoretically unrelated?}
%\cheng{Actually just "unrelated" is enough. Here I mean the query is like "What's the capital of the US", and the distraction is "The capital of China is Beijing". It's not related to what is inquired, but may still exert some interference, and there's the possibility that the model will answer "Beijing"}

We conclude by underscoring the inherent risks of hallucination and misinformation when introducing external knowledge, even inadvertently, that interferes with the LLM's parametric knowledge.

%Our study elucidates the manner in which LLMs elicit knowledge and manage potential conflicts, and underscores the inherent risks of hallucination when introducing external knowledge, even inadvertently, that interferes with the LLM's parametric knowledge.

\section{Related Work}

\paragraph{Internal and External Knowledge Conflicts.} LLMs amass internal knowledge through extensive learning on massive corpora during pre-training~\citep{roberts2020much, jiang2020can, gururangan2020don}, thereby weaving a unique system of parametric knowledge. This process, however, can be marred by inaccurate or outdated training data, leading to potential hallucinations within the model~\citep{carlini2021extracting, lazaridou2021mind, zhang2022counterfactual}. To align LLMs with current information and enhance factual accuracy, researchers have employed various tools~\citep{schick2023toolformer, qin2023tool}, memory techniques~\citep{zhong2022training}, and information retrieval strategies~\citep{guu2020realm, izacard2021leveraging}. However, such external knowledge may be novel or even contradict the model's existing parametric knowledge, causing interference. \citet{neeman2022disentqa} trained the model to disentangle internal and external knowledge and generate two responses to avoid conflict. \citet{zhou2023context} utilized special prompt engineering and abstention options to improve model faithfulness. More recently, \citet{xie2023adaptive} explored how the GPT model family reacts to knowledge conflicts, uncovering a high receptivity to external knowledge and confirmation bias. In line with these studies, we broaden our focus to encompass both black-box and open-source models, adopting a more systematic perspective on multiple types of distractors and parametric knowledge structures.

\paragraph{Propagation of Introduced Knowledge.} Prior approaches to model editing have primarily centered on the modification of parameters~\citep{meng2022mass, yao2022kformer} or the integration of specialized modules~\citep{wang2021k, mitchell2021fast} to enable the model to assimilate new knowledge. Nevertheless, this newly introduced external knowledge is anticipated to exert long-lasting effects. \citet{onoe2023can} have found that traditional editing methods exhibit inconsistencies when paraphrasing questions in new contexts, and that prepending entity definitions can facilitate the propagation of the injected external knowledge. \citet{zhong2023mquake} contributed a benchmark to measure how the alteration of one knowledge piece may influence the entire multi-hop QA chain's response. This phenomenon is termed the ripple effect by \citet{cohen2023evaluating}, who offer six evaluation metrics from this angle to assess the robustness of the model editing methods. Building on these investigations, we extend our focus to more hops and multiple knowledge structures, and build a generalizable framework to explore the impact of introduced external knowledge, particularly during the model's active engagement with users, and in a more controlled and systematic manner.

\begin{figure*}[!t]
    \centering
    \includegraphics[width=\linewidth]{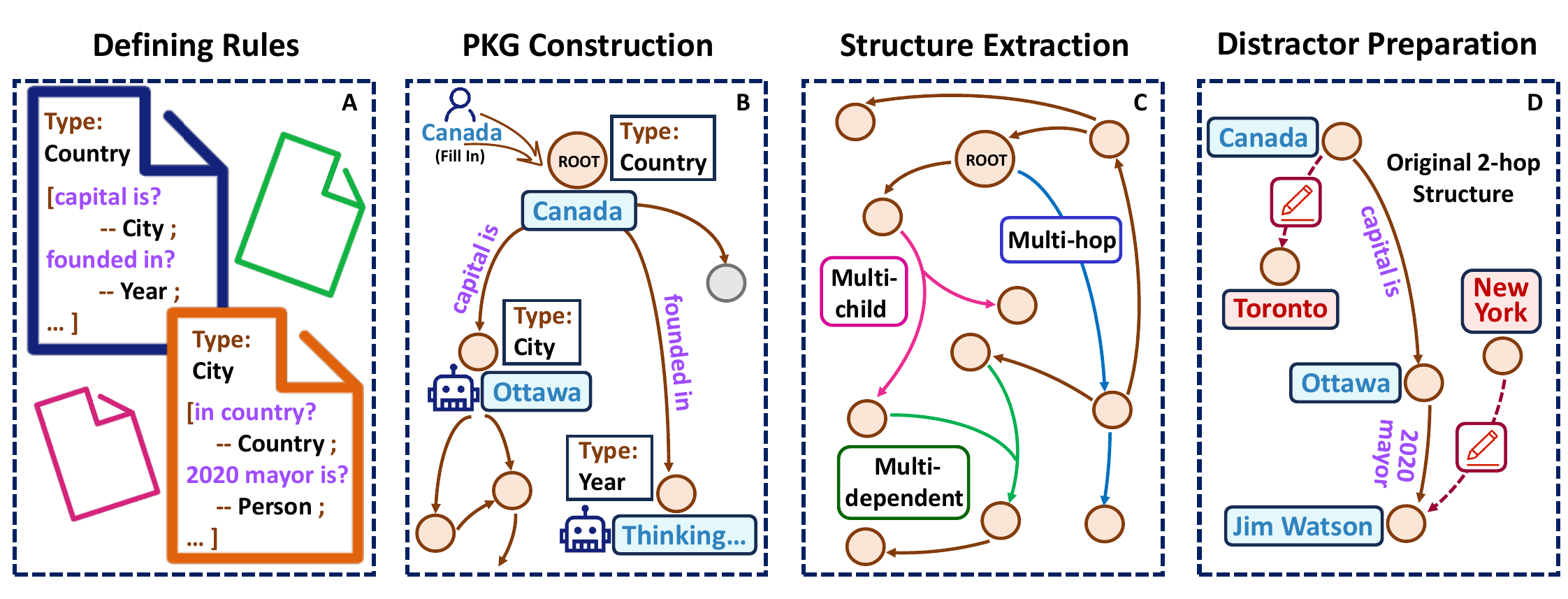}
    \caption{The pipeline for the construction of PKG and distractors. Figure A, B: The automatic construction of the model's PKG with defined rules. Figure C: The extraction of various PKG structures. Figure D: The modification of PKG to construct distractors.}
    \label{fig2}
\end{figure*}

\section{Introduction of Parametric and External Knowledge}

Our research question focuses on the impacts of \emph{external knowledge} on \emph{parametric knowledge}.
We discuss how we extract the parametric knowledge in a \emph{graph structure} to capture their relations, and introduce various types of external knowledge as \emph{distractors}.
%, and it naturally involves two aspects: revealing what the parametric knowledge is, and how the external knowledge should be introduced. 
%In this section, we introduce how we extract the model's parametric knowledge through the construction of the model's \textit{parametric knowledge }, and how we introduce the external knowledge through the creation of various types of \textit{distractors}.

\subsection{Parametric Knowledge Graph}
\label{Parametric_Knowledge_Graph}

LLMs have learned through pre-training a mass amount of parametric knowledge that is largely interconnected.
As mentioned in \cref{sec:intro}, these interconnections impact how LLMs react to different direct or indirect distractions in external knowledge. 
%To systematically investigate the model's response to external knowledge, we must first reveal its internal knowledge structure. 
To reveal such internal knowledge structure, we propose a novel framework to construct a model's \textit{parametric knowledge graph} (PKG).
%.leveraging the interconnected nature of the knowledge graph (KG).
% \sherry{Maybe add some detail on.} \cheng{I add some details later.}

PKG is a \emph{semantic net that integrates a model's parametric knowledge}.
It consists of nodes that represent \emph{entities} (denoted by $E$), and edges that represent \emph{relations} (denoted by $R$).
Drawing parallels to Knowledge Graph (KG), PKG allows for turning the \emph{implicit knowledge} within a model into \emph{explicit and structured} representations that are transparent for inspection and flexible for extension. Its \emph{automatic} construction also provides convenience for the extraction and modification of parametric knowledge.
% we \emph{automatically} organize knowledge \emph{in a structured way}, which allows for flexible extension and efficient inspection.
While traditional KGs are grounded in real-world facts~\citep{fensel2020introduction}, our approach stands as the first to use KG as an analogy for eliciting the model's parametric knowledge.

%\paragraph{Motivation} Our motivation for revealing the model's PKG involves multiple dimensions:
%\begin{itemize}[topsep=1pt, partopsep=1pt, leftmargin=12pt, itemsep=-3pt]
%\item \textbf{Transparency}: PKG turns implicit knowledge within a model into clear, explicit representation transparent for human investigation.
%\item \textbf{Extendability}: PKG allows for automatic construction and is easy to scale up.
%\item \textbf{Regularization}: The knowledge is organized in a structured way, which provides convenience for extraction and modification.
%\end{itemize}
%With these desired characteristics, PKG helps reduce the burden of automatic parametric knowledge inspection.

%\paragraph{Definition} PKG is a semantic net that integrates a model's parametric knowledge. It consists of nodes that represent \emph{entities}, and edges that represent \emph{relations}. Drawing parallels to KG, a PKG should be flexible to extend, support multiple relations, and permit efficient inspection.
%In addition, each entity in PKG possesses a core attribute \textit{type}, which governs the relations that it can extend. For instance, the type for the entity ``France'' would be ``Country''. The types and relations in a PKG are subject to customization.

\paragraph{Construction} 
To automatically construct the PKG, we provide a set of \emph{specification rules}, which defines what kinds of relations can exist in which types of entities. 
More concretely, we abstract each entity $E$ (e.g., France) in PKG into a \emph{type} (``Country''), and create rules for each type in the form of \texttt{($R$, \textit{target type})}.
For example, the entity type ``Country'' can extend the relation ``capital is'', targeting an answer of type ``City.''
% \sherry{Shall we color code the type, relation, and entity in the text according to their color in the figures? Just in these two paragraphs when we define things} \cheng{There's no explicit difference of colors in fig2 ..}
In practice, these rules are applied in natural language templates (\cref{Apdx_PKG_Construct}), which closely map the LLM's logic pathway to the graph in an interpretable way.
% \sherry{Provide an example of template?} \cheng{It's just we express the relation using natural language, illustrated in \cref{example_rules}. A bit trivial if explained in detail here.}

%PKG applies the natural language template in construction (please refer to Appendices~\ref{Apdx_PKG_Construct}), which makes the model's logic pathway more interpretable.

%The construction of PKG relies on a set of \sherry{specifically defined? well-defined?} specifically defined rules. The rules can be customized based on the attribute type. As shown in Figure~\ref{fig2}A, the relation for each type of entity assumes the form (\textit{relation}, \textit{target type}). For example, the entity type ``Country'' can extend the relation ``capital is'', targeting an answer of type ``City''.

Once the rules are set, a PKG can be automatically extended in a depth-first manner for any given root node (with an entity $E$ and its corresponding \emph{type}).
For example, in \cref{fig2}B, upon assigning the root node to ``Canada'', our framework sequentially extends all relations associated with the root type ``Country''. The model then seeks an answer corresponding to each target type, recursively shaping the whole PKG.
Experimentally, we apply consistency checks and only regard the answers that the model sticks to during consecutive queries as parametric knowledge (\cref{Apdx_PKG_Construct}).
% \sherry{how important is the root entity? Can we randomly pick one or does it have to be a perfect one?} \cheng{It could be a random one, but together with the max-depth defined, the choice of root will affect the magnitude of graph.}
% \sherry{Shall we define what we mean by parametric knowledge? top-1 answer to any question?} \cheng{Experimentally, it is the \textit{same} answer that model provides in three consecutive queries with different temperatures (consistency check, we mention in Appendix).}

%the user merely needs to specify the \emph{type} and \emph{entity} of the root node. This will trigger the model to automatically assemble the entire PKG in a depth-first manner. Illustrated in Figure~\ref{fig2}B, upon assigning the root node to ``Canada'', our framework sequentially extends all relations associated with the root type ``Country''. The model then seeks an answer corresponding to each target type, recursively shaping the whole PKG.

% ==================================== %

\begin{table*}[!t]
    \centering
    \includegraphics[width=\linewidth]{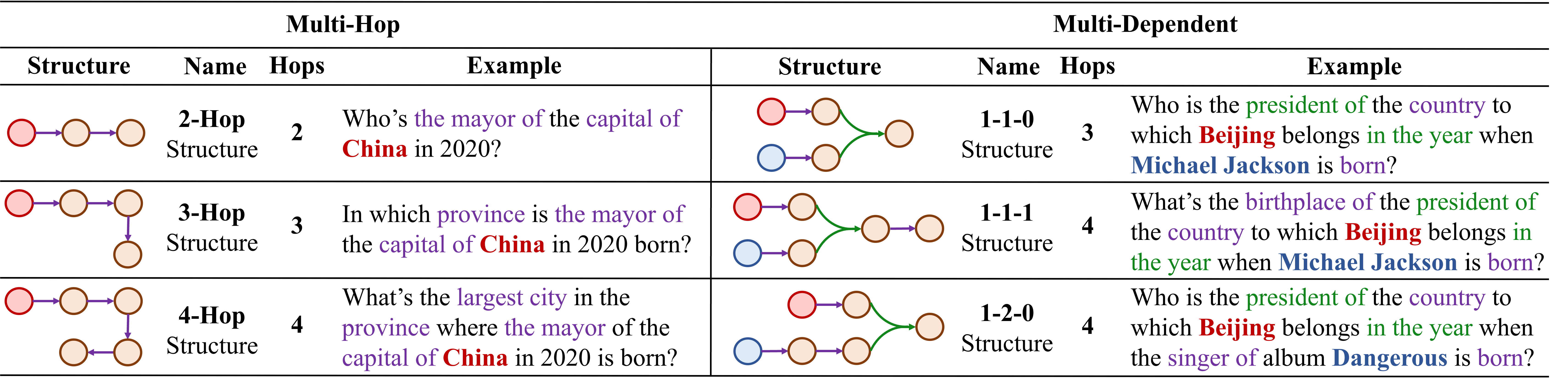}
        \caption{Three multi-hop and three multi-dependent structures we investigate in the experiments. The \deepred{red} and \blue{blue} nodes represent the starting entities, while other nodes are implicit (need reasoning to reach instead of directly given in the query \textit{Examples}). The \green{green} edges denote the multi-dependent relation, which is contained in the \textit{pivot hop}, while other \purple{purple} edges denote other explicit relations.} % \sherry{what's implicit node? Shall we define pivot hop here?} \cheng{Explained (later), subject to your further change.}}
    \label{structure_case}
\end{table*}

% ==================================== %

\paragraph{Extraction} 
One core advantage of PKG is that it enables us to \emph{extract data chains with structural variety}: As in \cref{fig2}C, PKG contains not only multi-hop structures, but also multi-child (where multiple answers exist for a given entity and relation) and multi-dependent relations (where two indispensable entities jointly decide the answer for a relation).
Such complexity enables the analyses of different types of relations in parametric knowledge.
% \textbf{Structural Variety}: The relations in PKG are not limited to triplet structures, which enables a more nuanced depiction of the model's parametric knowledge

% The complete PKG is too large for us to conduct a detailed analysis. \sherry{How large is too large?}
%This necessitates the extraction of various subgraphs as the basis for our study. As shown in Figure~\ref{fig2}C, PKG supports multi-child (where multiple answers exist for a given entity and relation) and multi-dependent relations (where two indispensable entities jointly decide the answer for a relation).

To support controlled experiments, we extract the sub-graphs from PKG with different structures, nodes, and edges.
To transform them into a usable format, we linearize each sub-graph into a ``data chain'' represented by triplets $[(E_0, R_1, E_1), (E_1', R_2, E_2), ..., (E_{n-1}', R_n, E_n)]$ ($E_k = E_k'$ for multi-hop chains).
As shown in \cref{structure_case}, we mainly apply the multi-hop (2, 3, and 4-hop) structures as the basis of queries in experiments;
To further capture the non-linearity, we define three \emph{multi-dependent structures}~\cite{trivedi2022musique}, each containing multiple hops and a core multi-dependent relation (\cref{structure_case}, right).
We denote the hop that contains the multi-dependent relation in the data chain as the \emph{pivot hop}.
% \sherry{I still don't know what's a pivot hop?} \cheng{I think we still use "multi-dependent", or change to something that still emphasizes "multi", instead of only 1 parent as the upstream dependency.} \cheng{Pivot hop only refers to the hop of query that contains multi-dependent relation (green color in Table 1, right column). We differentiate structure (a sub-graph / data chain with multiple hops), hop (a query [entity, relation], and an answer [entity]), and relation (edge in PKG) here: Each multi-dependent structure must contain one multi-dependent relation, this relation is in the pivot hop of the data chain.}

The answer for the pivot hop relies on two upstream entities, and both of them can serve as the ending node for multi-hop chains of lengths $A$ and $B$, respectively. Simultaneously, the answer entity for the pivot hop can serve as the starting node for a multi-hop chain of length $C$. The values of $A$, $B$, and $C$ collectively govern the specific configuration of the multi-dependent structure, succinctly referred to as the $A$-$B$-$C$ structure.

\begin{figure*}[!t]
    \centering
    \includegraphics[width=\linewidth]{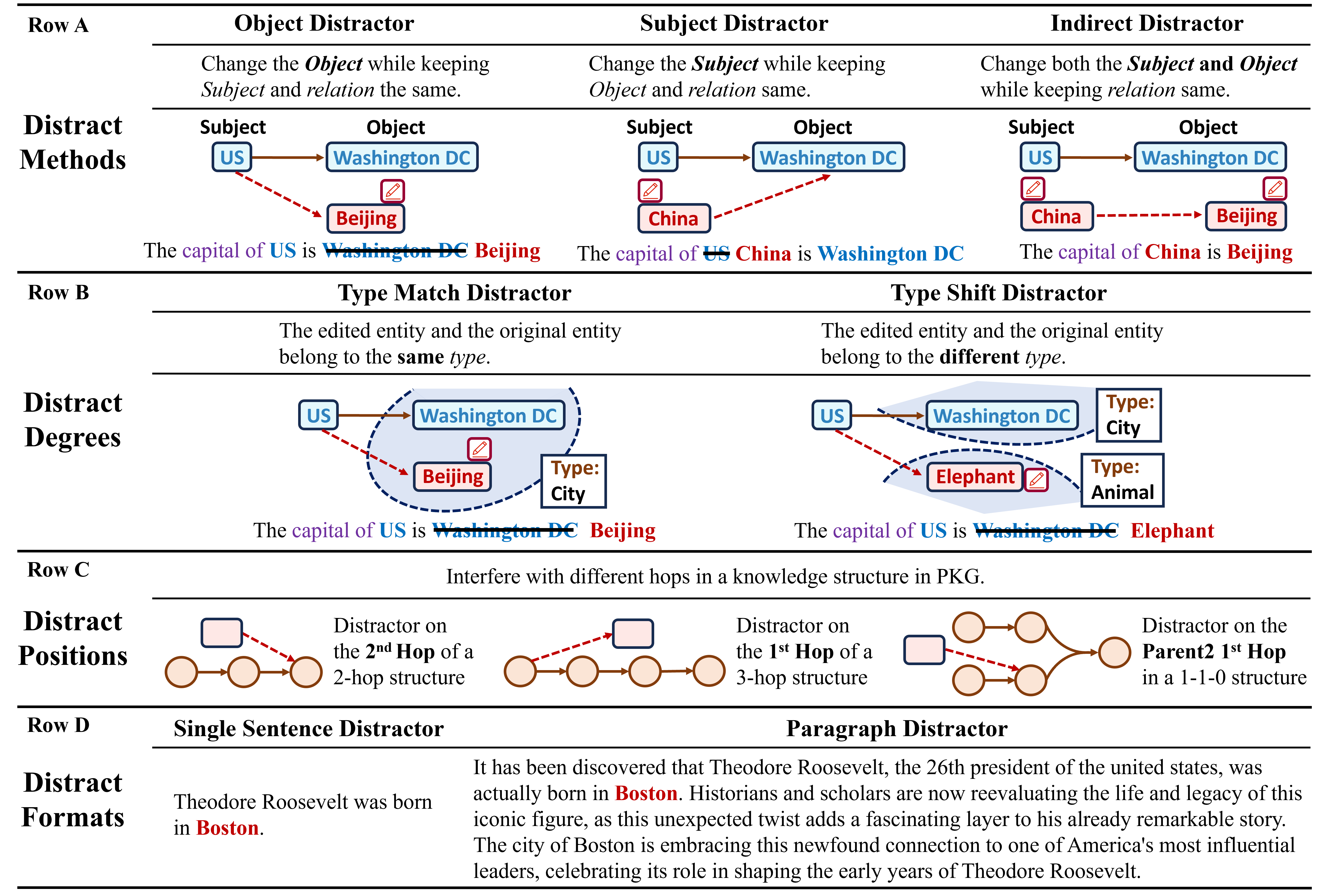}
    \caption{An Illustration of different types of distractors we apply in experiments.}
    \label{fig3}
\end{figure*}

% ==================================== %

\subsection{External Knowledge Distractors}

The external knowledge in our experiments is introduced through \textit{distractors}, which serve as the source of interference to the model's PKG.
% Without distractors, the model provides the same output as the original PKG in most cases. \sherry{? why not always? And do we need this sentence here} \cheng{I state this as you have commented last time lol, but feel like we don't really need it here.}

Distractors are directly derived by modifying the extracted raw data chains from the model's PKG. \cref{fig2}D shows a simplified example where the capital of ``Canada'' is substituted with ``Toronto'', and the subject of ``2020 mayor'' is replaced with ``New York''. In both cases, distractors are derived from modifying a 2-hop chain.
We prompt GPT-3.5 for the distractor's automatic construction, as detailed in \cref{Apdx_Distractor_Construct} and \cref{prompt_object_match} to \cref{prompt_indirect_shift}. The distractor will be presented as a natural language description at the beginning of model-user interaction shown in \cref{fig1}.
% \sherry{In prompts, distractors are constructed using GPT-3.5, and presented as a natural language description before... (\cref{fig1}). More details in \ref{Apdx_Distractor_Construct}. --- In general I think we should connect the framework more closely to how they map to the prompts, to provide more reproducibility} \cheng{I add a sentence here. This is originally arranged in the description of experiment methods, in Sec.4.1.}

We systematically create distractors by varying four dimensions (\cref{fig3}): distract methods, degrees, positions, and formats, as detailed below.
These types of distractors allow for a nuanced exploration of how alterations in different components of the PKG can lead to varied effects in the model's responses.

%To investigate in a more systematic way, we divide the distractors into different types based on their attributes, namely distract degrees, methods, positions, and formats. \cref{fig3} shows an overview for all the variations.

\paragraph{Distract Methods} Different distract methods reveal how the external knowledge is related to the original parametric knowledge. 
As in \cref{fig3}A, \textit{Object Distractor} introduces distraction by changing the object of the original parametric knowledge in the raw data chain. For instance, by changing the original object ``Washington DC'' into ``Beijing'' while preserving the subject and relation, the resulting external knowledge ``The capital of US is Beijing'' constitutes an \textit{Object Distractor}. \textit{Object Distractor} often represents an explicit contradiction to the model's original parametric belief.
% \sherry{Either here or in 3.1 (maybe in extraction?) we should clarify what questions we are asking LLMs, because this changes why object vs. subject makes a difference. At a higher level, when we describe these dimensions, we should highlight why we think these are impactful.} \cheng{Added justification to each paragraph's ending.}

Similarly, \textit{Subject Distractor} introduces distraction by changing the subject, while \textit{Indirect Distractor} changes both the subject and the object while preserving the relation.
As we always query for the \textit{Object} in knowledge given the \textit{Subject} and \textit{Relation}, this makes \textit{Object Distractor} an explicit contradiction to the model's original parametric belief, while the other two are ``weaker''.
This distinction may lead to different resulting impacts.

\paragraph{Distract Degrees} Various distract degrees illustrate how severely the external knowledge deviates from the original parametric knowledge (\cref{fig3}B). 
This deviation is measured through \emph{type}: We define the distractor as \textit{Type Match} if the edited entity and original entity belong to the same type (e.g., city ``Washington DC'' to city ``Beijing''), and \emph{Type Shift} if otherwise (e.g., city ``Washington DC'' to animal ``Elephant.'')
Because of the change in the underlying type, \textit{Type Shift Distrcators} are often evidently false information, while \textit{Type Match Distrcators} are more confounding to models. This makes distract degrees important as they reflect how well could the models accept knowledge with different credibility.
% \sherry{can we assign type to entities? can elephant be "city"?} \cheng{No, it is based on rules and is the intrinsic property of the entity. During the user-model interaction, \textit{type} is implicit.}
% For instance, changing ``Washington DC'' to ``Beijing'' results in a \textit{Type Match Distractor} as both entities are cities. However, if the resulting edited entity is ``Elephant'', then this constitutes a \textit{Type Shift Distractor} as the elephant belongs to the type animal.

% We note that methods and degrees provide distinct perspectives for describing a distractor. Therefore, they are mutually unrelated, and one distractor can possess both attributes (e.g. in Figure~\ref{fig3}, an \textit{Object Distractor} can also be a \textit{Type Match Distractor}.). By combining three different methods and two different degrees, a total of six types of modifications can be applied to a piece of original parametric knowledge, each turning into a unique type of distractor.\sherry{feels like we don't need this paragraph, because a distractor can easily have any of these four attributes}

\paragraph{Distract Positions} Built upon distract methods and degrees, different distract positions reflect to which relation in the extracted data chain is the external information introduced. This attribute of the distractors doesn't describe how to concretely modify the knowledge into distractions, but rather where to introduce this distraction.
In \cref{fig3}C, we present three examples of different distract positions. The total number of positions that external knowledge could be introduced is decided by the total number of hops in a knowledge structure.
Different distract positions in essence represent different stages in the evolving user-model interaction. For multi-dependent structures, we can also utilize distract positions as a means to distinguish the unique impact of introducing distractions to the pivot hop.
%as we have listed in Table~\ref{structure_case}.
%In any hop of a data chain, 6 types of distractors (3 methods $\times$ 2 degrees) can be introduced.

\paragraph{Distract Format} The distract format differentiates the context of external knowledge. 
%It also builds upon distract methods and degrees. 
Based on the context length, we introduce \textit{Single Sentence Distractor}, which states the external knowledge in one simple sentence, and \textit{Paragraph Distractor}, which describes the knowledge through 3-4 sentences with supporting details. In \textit{Row D} of Figure~\ref{fig3}, we illustrate how a simple piece of external knowledge ``Theodore Roosevelt was born in Boston'' can be extended to a paragraph.
Different distract formats are introduced to prove whether the model possesses a certain bias towards lengthier and more detailed descriptions.

%\paragraph{Significance} All these types of distractors allow for a nuanced exploration of how alterations in different components of the PKG can lead to varied effects in the model's responses. For each type of distractor, we also employ GPT3.5 for automatic construction. Please refer to Appendices~\ref{Apdx_Distractor_Construct} for the prompts and more details.

% ==================================== %

\begin{figure}[!t]
    \centering
    \includegraphics[width=\linewidth]{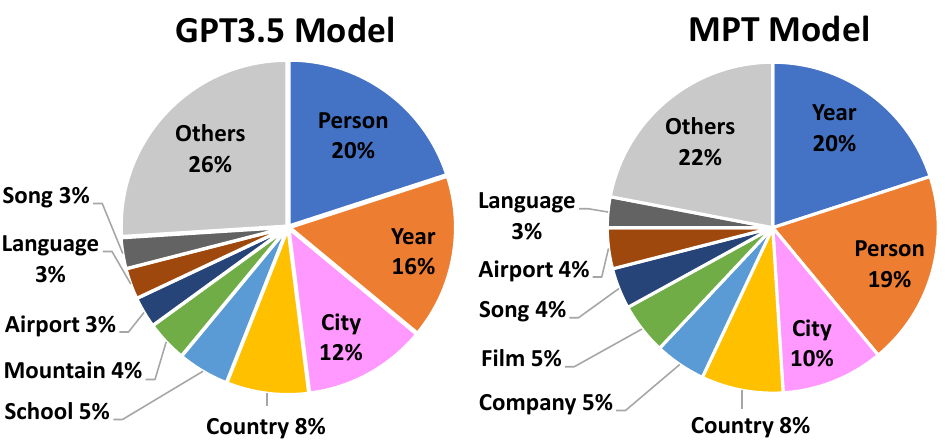}
    \caption{Ratio of different \textit{types} in model's PKG.}
    \label{type_stat}
\end{figure}

\begin{table}[!t]
\centering
\small
\begin{tabular}{r| r | r | r}
\toprule
%\multirow{2}{*}{\textbf{Rules}}
%& Total Types & \multicolumn{2}{l}{17} \\
%& Total Rels & \multicolumn{2}{l}{63} \\
\multicolumn{2}{r|}{Dimensions / Model} & \textbf{GPT3.5} & \textbf{MPT-7B} \\
\toprule
\multicolumn{2}{r|}{Avg Node Num}&      278      &      166     \\
\midrule
\multicolumn{2}{r|}{Avg Edge Num}&      467      &      276     \\
\midrule
\multicolumn{2}{r|}{Multi-dependent Rels}&      769      &      443     \\
\midrule
\multicolumn{2}{r|}{Multi-child Rels}&      192      &      124     \\
\midrule
\multirow{3}{*}{\makecell{Multi-hop \\ Structures}}
& 2-hop  &      5,361      &      3,360     \\
& 3-hop  &      14,523     &      8,642     \\
& 4-hop  &      28,297      &      17,064     \\
\bottomrule
\end{tabular}
\caption{The statistics of 8 PKGs we apply respectively for GPT3.5 and MPT-7B.
\textit{Rels} denotes relations. On average, GPT3.5 exhibits a more expansive and intricate PKG. The magnitude of distinct relations and varied structures within the PKGs exemplifies their heightened diversity and complexity.}
\label{stat_findings}
\end{table}

% ==================================== %

\section{Experiment Setup}
To understand the model's reaction when external knowledge interferes with its parametric knowledge, we conduct experiments by systematically varying the combinations of the model's parametric knowledge structures and the external knowledge distractors.
% \sherry{Add a sentence on why we are running the experiment and "to this end we vary combinations of internal and external..."} \cheng{Added.}

\subsection{Method}
Each structure we extract from PKG contains multiple hops of queries, which constitute the data chain that represents the model's original logic pathway. As we aim to inspect the model's responses during active interaction instead of testing its multi-hop reasoning ability, we follow the ``instance-wise'' probing method proposed by \citet{zhong2023mquake} and test the data chain in a one-hop manner after the introduction of distractors.

As illustrated previously in Figure~\ref{fig1}, we first present the distractor to the model as the introduced external knowledge. Next, for the data chain $[(E_0, R_1, E_1), (E_1', R_2, E_2), ..., (E_{n-1}', R_n, E_n)]$ extracted, where two entities $E$ and a relation $R$ form a knowledge triplet, we first probe for the model's answer $A_1$ after given $(E_0, R_1)$. Then, we continue to probe the next hop of the query, while giving the model all previous interaction history. The new query is based on $(A_1, R_2)$ if $E_1' = E_1$ (this always holds for multi-hop structures) or based on $(E_1', R_2)$ if $E_1' \neq E_1$ (this only happens for multi-dependent structures), and we ask for the model's answer $A_2$. This iterates until all the queries are done or the model abstains from answering.
% \sherry{add these notions in sec 3 when you talk about entities relations and rules} \cheng{Added, and we state in a more specified way.}

\paragraph{Controlled Settings}
To control the variables in our experiment, for all the studies except knowledge structures, we experiment on all the \textit{multi-hop} structures as raw data chains. For all the studies except the external knowledge format, we apply \textit{Single Sentence} as the distractor's knowledge format. Please refer to \cref{Apdx_Exp_Setting} for a more detailed explanation of each experimental setting.
% \sherry{I moved this here -- makes sense?} \cheng{Looks Good!}

\subsection{Models}
% \sherry{add link and citation for the models?} \cheng{Added}
We conduct experiments utilizing both the open-source MPT-7B~\citep{mosaic2023mpt} and the black-box GPT3.5 models~\citep{openai2022chatgpt}. MPT-7B and GPT3.5 are selected for their robust interaction capabilities with users, which aligns well with our experimental design. Furthermore, the open-source nature of MPT-7B enables the analysis of confidence values. Please also refer to \cref{Apdx_Exp_Setting} for more details about the hyper-parameters we apply. We also present some experiment results from GPT3 as additional support to our findings in \cref{Apdx_Exp_Davinci}.

\subsection{Data}
\label{Setup_Data}
We construct 8 PKGs with different root nodes for both GPT3.5 and MPT-7B using manually defined rules. In total, we used 17 types and 63 relations in the construction rules. The statistical findings of the raw PKGs we apply are summarized in Figure~\ref{type_stat} and Table~\ref{stat_findings}.
% \sherry{ In total, we used 17 types and 63 relations in the construction rules. --- moved this to here} \cheng{Moved}
For all the studies besides the knowledge structures in PKG, we employ N-hop data chains ($N \in {2, 3, 4}$), utilizing 200 chains for each type. Each $N$-hop data chain affords $N$ positions for external knowledge introduction, three distract methods, and two distract degrees, resulting in $6N$ rounds of queries (or, $6N$ different distractors) and $6N^2$ hops of queries per original chain. Consequently, these constitute 10,800 query rounds, encompassing a total of 34,800 query hops. 

For the study on knowledge structures in PKG, we extract 100 raw data chains for each multi-dependent structure type illustrated in the right column of Figure~\ref{structure_case}. The collected chains constitute 6,600 query rounds, encompassing a total of 24,600 query hops. The tool for automatic PKG construction and all the data we apply is released.

\subsection{Metrics}

\paragraph{Consistency}
Our primary metric is \emph{consistency}, which quantifies whether the model will stick to the answer in its PKG during multiple rounds of queries even with the presence of distractors.
Formally, among $N$ query chains $C_1, C_2, ..., C_N$, the model outputs the \emph{final answer} that adheres to its PKG in $M$ chains. \textit{Consistency} is defined as:
$$\text{Consistency}(\{C_1, ..., C_N\}) = \frac{M}{N}.$$
It reflects the ratio of query chains from PKG that are not affected by the external distractors.

\paragraph{Confidence} Moreover, inspired by \citet{kadavath2022language}, we also explore the MPT-7B's likelihood of outputting the target entity through the computation of \textit{confidence}.
Given the tokens ${t_0, t_1, ..., t_M}$ of a core entity $E$ in the model's response, the model's confidence in outputting this entity as the answer is defined as:
$$\text{Confidence}(E) = \prod_{t=t_0}^{t_M} \frac{e^{z_t}}{\sum_{i=1}^{N} e^{z_i}},$$
where $z_t$ denotes the raw score (logit) associated with the token $t$, and $N$ denotes the total number of tokens in the vocabulary.

Diving deeper into the model's responses, we further look into concrete model behaviors both at the final answer and across the chain.
First, for the \textbf{final answers}, we split \emph{inconsistent} answers into \emph{abstention} (when a full query chain cannot be completed because the model starts to answer e.g., ``I don't know'' at certain hops in a chain), and \emph{variation} (when the model reaches a final answer, but not same as the original PKG).
Then, \textbf{across the chain}, we also study if the model's answer for a particular hop of the query adheres to its PKG or not by categorizing each model's responses as either \emph{conforming} (when the model answers as its original PKG) or \emph{deviated} (when the model's answer is from the distractor).

% ========== all results table =========== %

\begin{table*}[!t]
    \centering
    \includegraphics[width=1.0\linewidth]{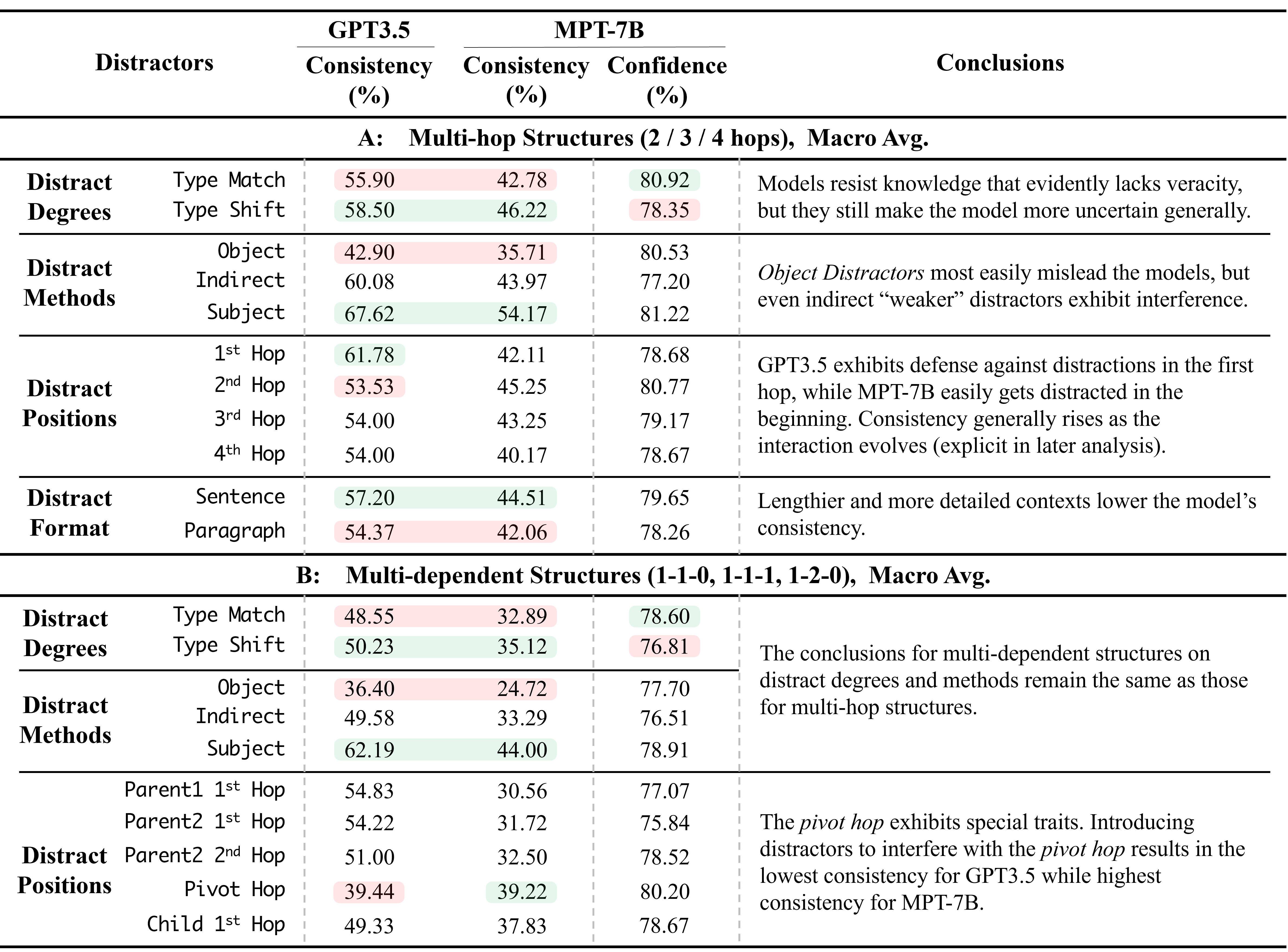}
    \caption{An overview of our experimental results and conclusions. We provide results on GPT3.5 and MPT-7B's consistency and confidence (macro-average on different structures), investigating the impacts of various distractors on multi-hop and multi-dependent structures in PKG. Results that we focus on are shaded: \sethlcolor{shadegreen}\hl{green} is applied for the \textit{highest} numerical value for a distractor type, while \sethlcolor{shadered}\hl{red} is applied for the \textit{lowest}. Please refer to the Appendix for detailed scores of different structure types instead of macro-average.} % \sherry{make the \emph{distractor names} right aligned, and try \\texttt e.g., something like Monaco? } \cheng{Is this better? I also change to \\texttt in \cref{method_conf_deviate}. If so I would change to \\texttt to all others.} \sherry{looks good!!}}
    \label{results_all_avg}
\end{table*}

% ==================================== %

% ==================================== %

\begin{figure}[!t]
    \centering
    \includegraphics[width=\linewidth]{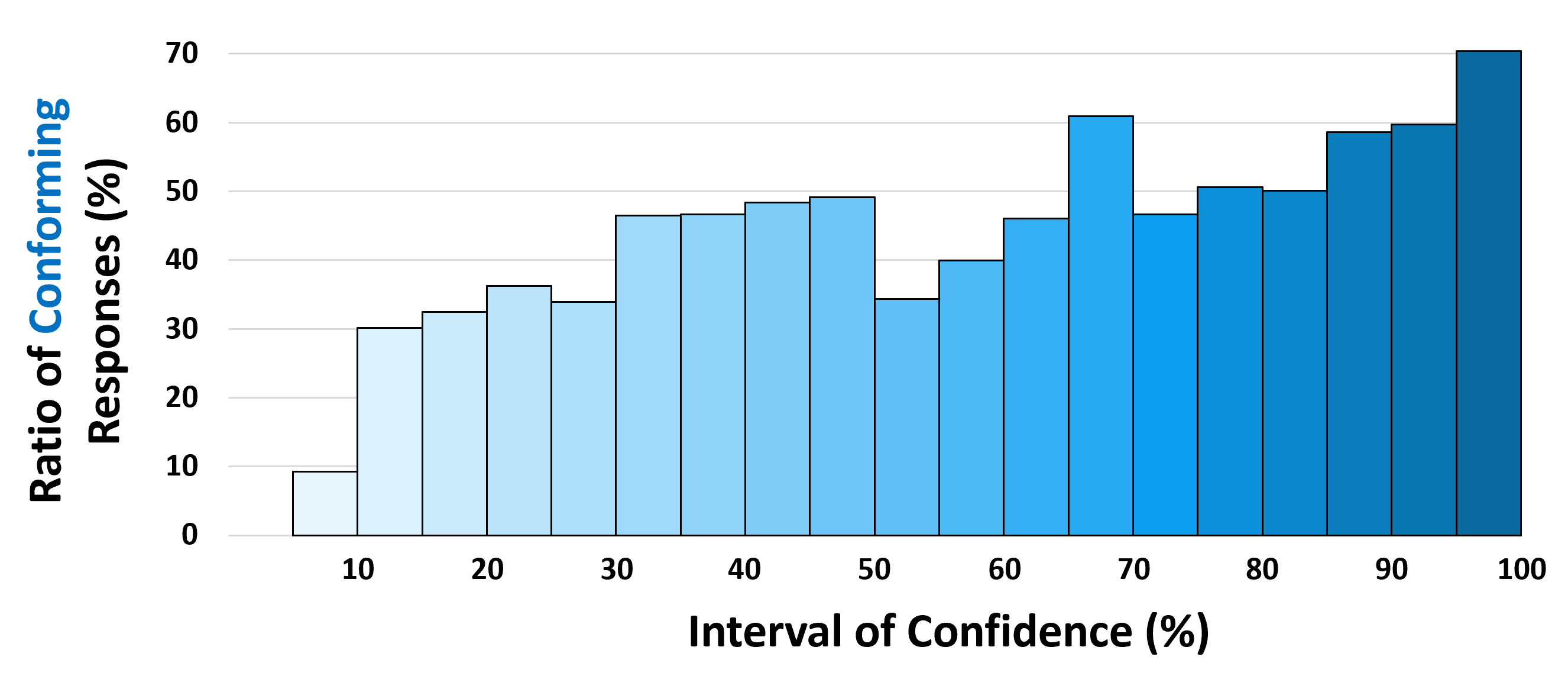}
    \caption{Statistics of the ratio of conforming responses with respect to the confidence placed in the corresponding relations in PKG. From left to right, as the confidence in a particular relation in PKG rises, the model is more likely to provide answers that conform with the PKG despite the presence of distractions.}
    \label{pilot_conf_analysis1}
\end{figure}

\begin{figure}[!t]
    \centering
    \includegraphics[width=\linewidth]{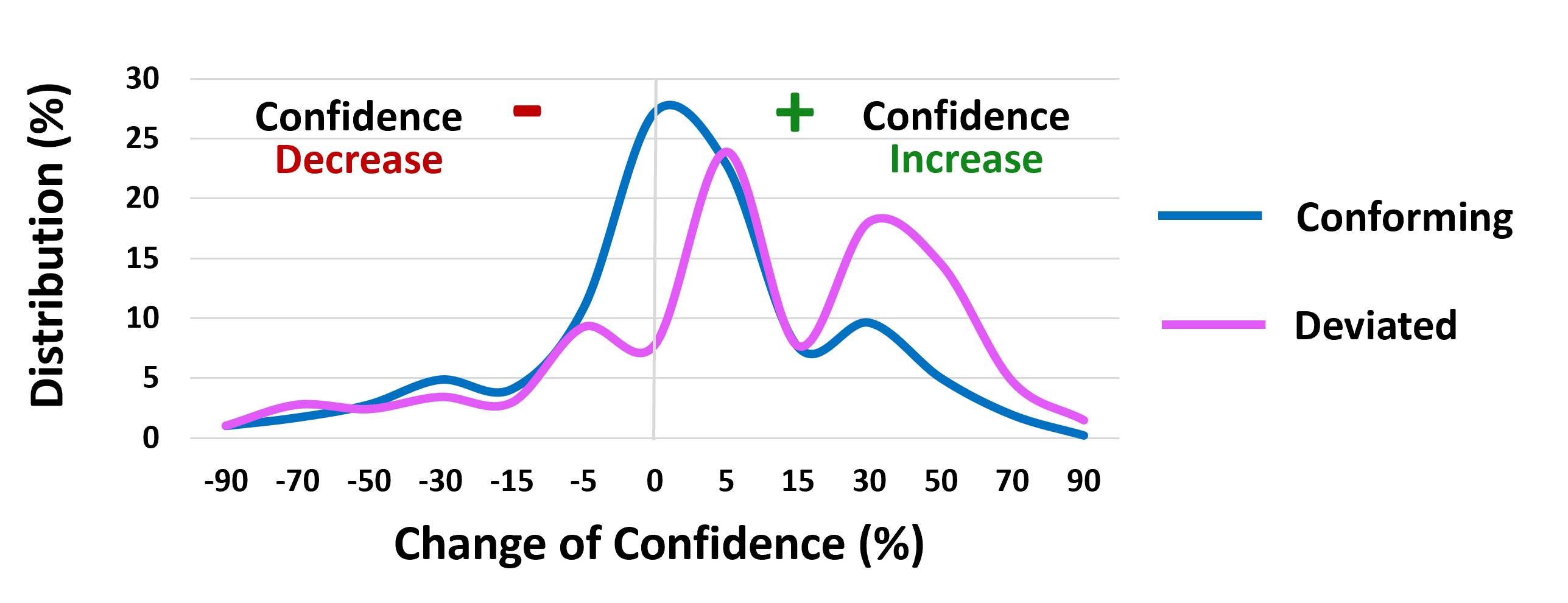}
    \caption{The distribution of the \textit{change} of confidence after introducing the distractors with respect to conforming and deviated responses. The area under the curve left of 0 represents the ratio of negative confidence change, and vice versa. With external knowledge, the model's confidence generally increases, especially for deviated responses.}
    \label{pilot_conf_analysis2}
\end{figure}

% ==================================== %

\section{Experiment Results}

\subsection{Effectiveness of Distractors through Confidence Analysis}
\label{Pilot_Studies}

Before diving into the specific effects of different distractors, we first analyze in general \textit{why the distractors we introduce are effective}. We conduct analysis through the lens of confidence to unveil the mechanism behind the model's responses under distractions.

\paragraph{Consistency occurs with high confidence.} We observe that the model is more likely to provide responses conforming to PKG when it is already confident with this piece of knowledge in PKG. In Figure~\ref{pilot_conf_analysis1}, we show the ratio of conforming responses generally rises as the confidence of that queried relation in the model's PKG increases. From another perspective, this also indicates if the model lacks confidence in a particular knowledge from PKG initially, the distractor is more likely to succeed in causing deviation during later queries.
% \sherry{personally I think paragraph title can be written as a natural sentence and no need to be capitalized.}

% ===== start on the flight =====
\paragraph{Response deviates with raised confidence.} We find that the model's confidence generally \emph{rises} with the introduction of external knowledge, especially for the deviated responses. In Figure~\ref{pilot_conf_analysis2}, we measure how the confidence \textit{changes} with the introduction of distractors. We plot the distribution of changes respectively for the conforming and deviated responses, and discover that: i) The area under positive confidence change is generally larger, indicating that external knowledge in general bolsters the model's confidence. ii) Most of the deviated responses experience an increase in confidence, proving that the distractors we introduce can deviate the model's responses with generally higher confidence.

\subsection{Results on Different Distractor Types}
After gaining a general understanding of how distractors are effective through their interactions with the model's confidence, we continue to investigate the impacts of distract degrees, methods, positions, and knowledge formats. The numerical results and major conclusions are presented in Table~\ref{results_all_avg}A.

% ==================================== %

\begin{table}[!t]
    \centering
    \includegraphics[width=\linewidth]{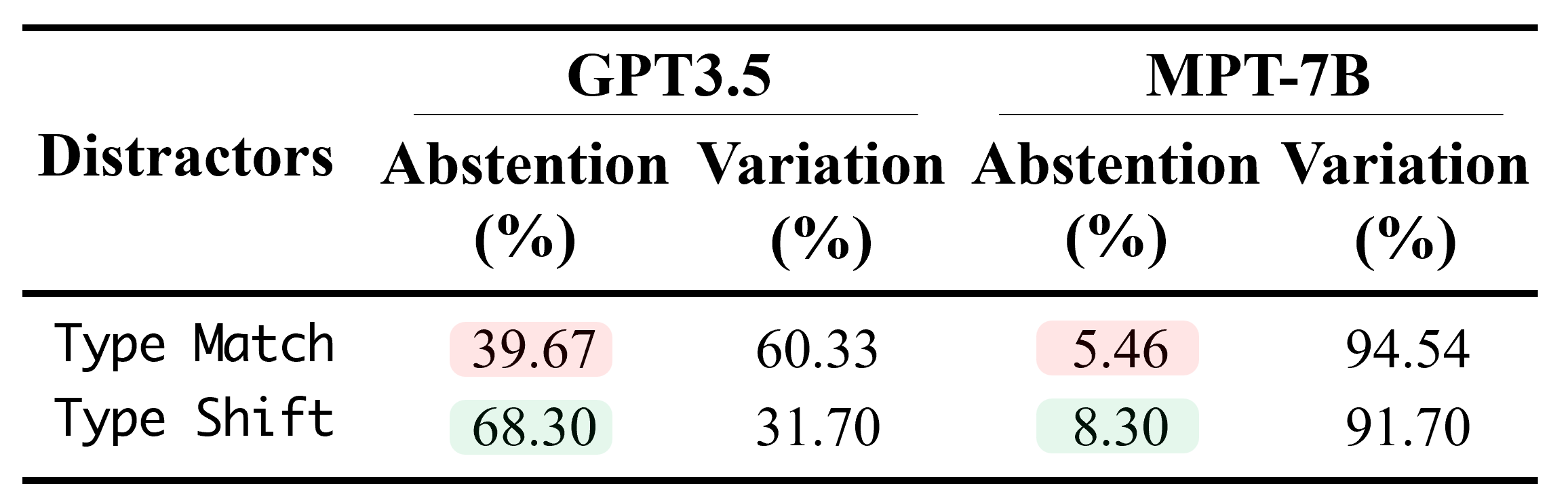}
    \caption{The rate of abstention and variation in the inconsistent chains when confronting distractors of different degrees for both GPT3.5 and MPT-7B. The results are the macro-averages on three multi-hop structures. \textit{Type Shift Distractors} cause more abstentions.}
    \label{degree_error}
\end{table}

% ==================================== %

\paragraph{Distract Degrees: Models exhibit resistance to knowledge that evidently lacks veracity.} We discover that compared to \textit{Type Match Distractors}, \textit{Type Shift Distractors} are less successful in misleading the model's responses. In the first row of Table~\ref{results_all_avg}A, we show the consistency is higher on \textit{Type Shift Distractors} for both GPT3.5 and MPT-7B. The observations are significant: With a Student's t-test, we obtain $p<0.001$ in both cases (\cref{Apdx_Distrcat_Degrees}). As \textit{Type Shift Distractors} changes the type of the edited entity and often yields external knowledge beyond commonsense (e.g. ``The capital of US is \textit{Elephant}''), our results demonstrate the LLMs are resistant to such knowledge that obviously lacks veracity.

Nevertheless, the overall confidence of the model's responses decreases for \textit{Type Shift Distractors}, suggesting that while the model may reject such distractors, the presence of these severely altered information can still exert strong effects of uncertainty.

We also investigate the inconsistent chains and discover that: i) Compared to MPT-7B, GPT3.5 is more likely to abstain under interference. ii) Compared to \textit{Type Match Distractors}, \textit{Type Shift Distractors} are more likely to cause abstention. The error analysis is presented in Table~\ref{degree_error}. We provide more detailed results and further analysis to confidence in \cref{Apdx_Distrcat_Degrees}.

% ==================================== %

\begin{table}[!t]
    \centering
    \includegraphics[width=\linewidth]{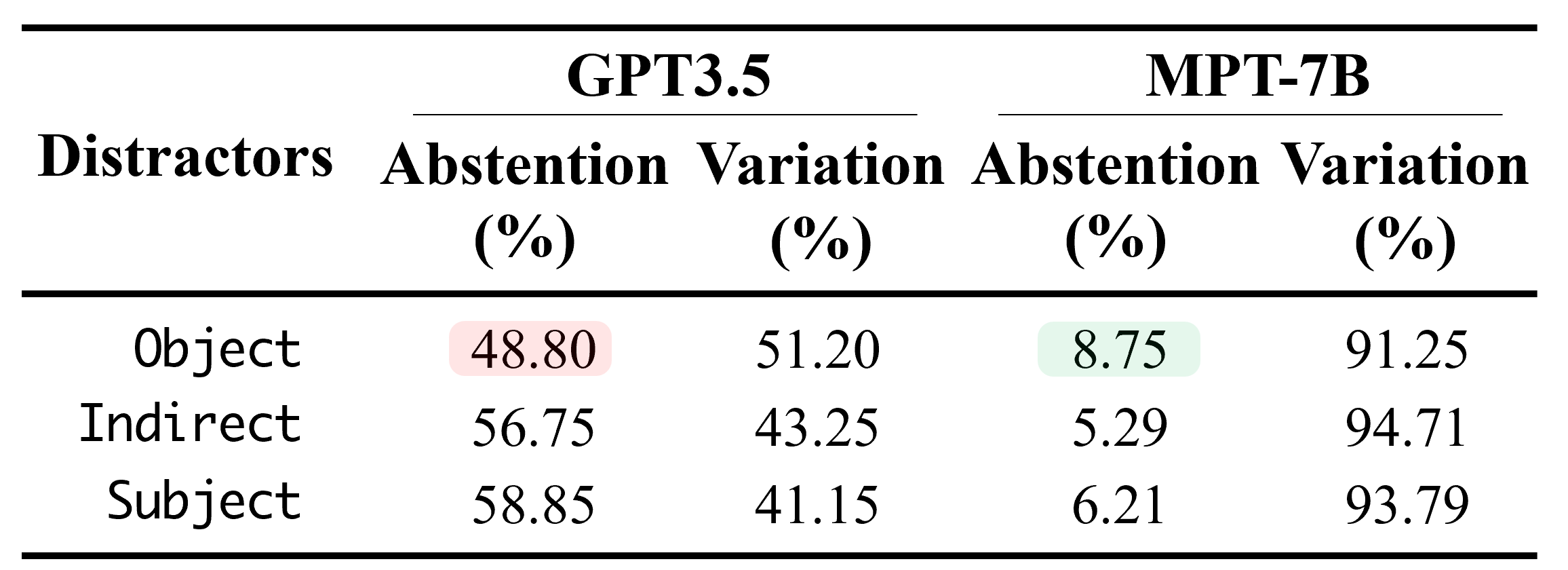}
    \caption{The rate of abstention and variation in the inconsistent chains when confronting distractors applying different methods for both GPT3.5 and MPT-7B. The results are the macro-averages on three multi-hop structures. \textit{Object Distractors} cause the least abstention in GPT3.5, while the most for MPT-7B.}
    \label{method_error}
\end{table}

% ==================================== %

\paragraph{Distract Methods: \textit{Object Distractors} lead to the lowest consistency, while \textit{Indirect Distractors} also exhibit interfering effects.} We observe that \textit{Object Distractors}, which bring directly conflicting external knowledge, particularly drive the model's deviation from its PKG in responses. As presented in the second row of Table~\ref{results_all_avg}A, consistency is the lowest for \textit{Object Distractors} among the three methods for both GPT3.5 and MPT-7B (P-value $p<0.001$ in all cases with Student's t-test, detailed in \cref{Apdx_Distrcat_Methods}). This could be attributed to the fact that only \textit{Object Distractors} retain the subject of original knowledge, thus forging an erroneous relation link that diverges from the model's original logic pathway. This more easily leads the model to a false final answer.
% \sherry{when p value is smaller than 0.001 just say <0.001}

For inconsistent chains, we discover that \textit{Object Distractors} lead to the lowest abstention in GPT3.5 but the highest in MPT-7B. This indicates that among the three distract methods, GPT3.5 is less likely to abstain if the distractor is a direct conflict, while MPT-7B shows vice versa. The error analysis is presented in Table~\ref{degree_error}. We also provide more detailed results and further analysis to confidence in \cref{Apdx_Distrcat_Methods}.

% ==================================== %

\begin{table}[!t]
\centering
\small
\tabcolsep=0.032\linewidth
\begin{tabular*}{\linewidth}{cccc}
\toprule
\textbf{Distractors} 
& \makecell{\texttt{Object}} & \makecell{\texttt{Indirect}} & \makecell{\texttt{Subject}} \\
\toprule
\textbf{Confidence (\%)}
&  77.70  &  74.31  &  71.90  \\
\bottomrule
\end{tabular*}
\caption{The average confidence of relations in the PKG that the model deviates in later responses. ``Weaker'' distract methods tend to mislead the model on the knowledge that it is originally not confident about.}
\label{method_conf_deviate}
\end{table}

\begin{figure}[!t]
    \centering
    \includegraphics[width=\linewidth]{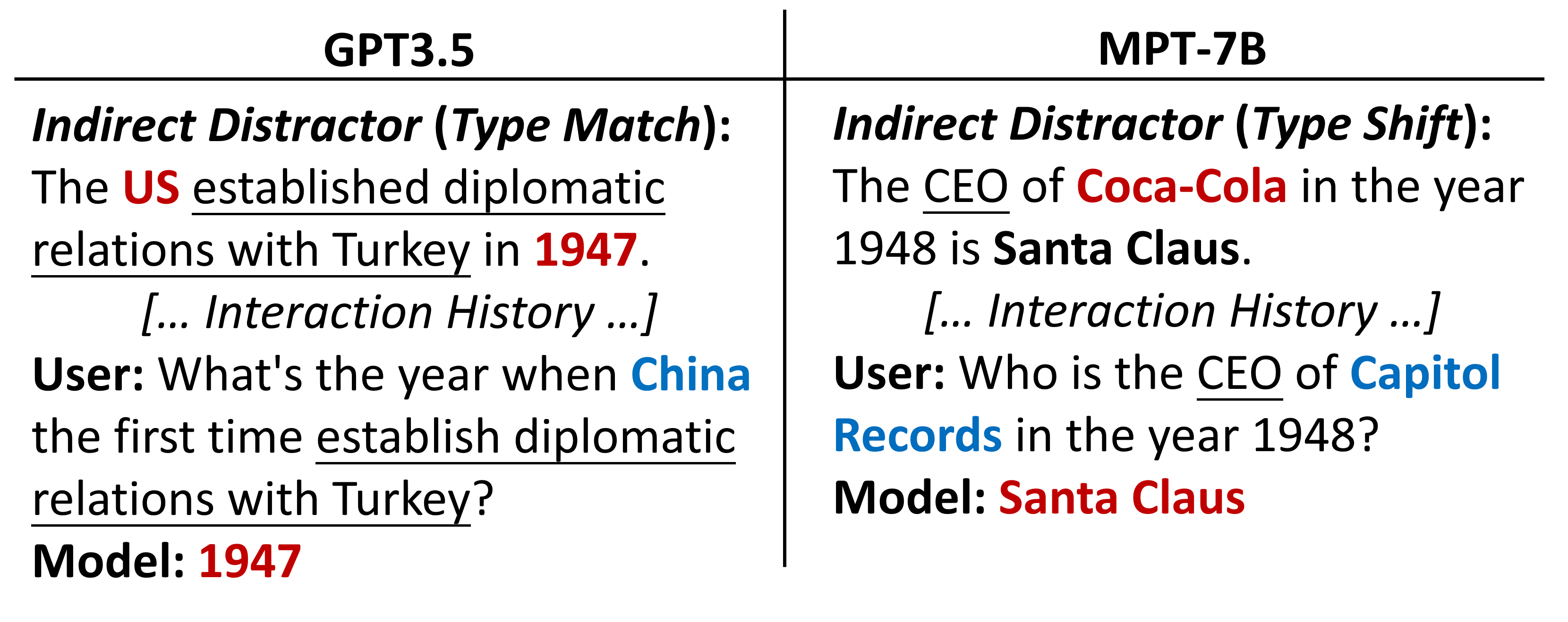}
    \caption{Case study on how the model deviates in response under \textit{Indirect Distractors}. The weak belief of target knowledge in its PKG and some intrinsic similarity in details mislead both models.}
    \label{method_cases}
\end{figure}

% ==================================== %

Despite the direct misleading effect of \textit{Object Distractors}, we observe \textit{Subject} and \textit{Indirect Distractors} also exhibit interference. Both methods alter the subject and make the introduced external knowledge not explicitly align with the query, so they are ``weaker'' distract methods that should not directly mislead the model. Why these ``weaker'' distractors also lead to low consistency?

Inspired by previous studies in \cref{Pilot_Studies}, we analyze the confidence of relations within the PKG that the model subsequently deviates in its responses. We discover when confronting a ``weaker'' distract method, the model is more susceptible to deviate in queries where its initial belief in the PKG is also weak. As showcased in Table~\ref{method_conf_deviate}, the average confidence of relations that later lead to deviations is the lowest for \textit{Subject Distractors}, followed by the \textit{Indirect Distractors}.

Specifically, we present a case study in Figure~\ref{method_cases}: For GPT3.5, the uncertainty about the user's query drives it to extract ``1947'' from the distractor as the final answer, despite the subject in distractor being ``US'' rather than ``China''` as inquired. The same happens to MPT-7B, as the same additional information ``in the year 1948'' presented in both the query and the distractor drives the model to trust ``Santa Claus'' as the company's CEO, though it is evidently false.

% ==================================== %

\begin{figure}[!t]
    \centering
    \includegraphics[width=1.0\linewidth]{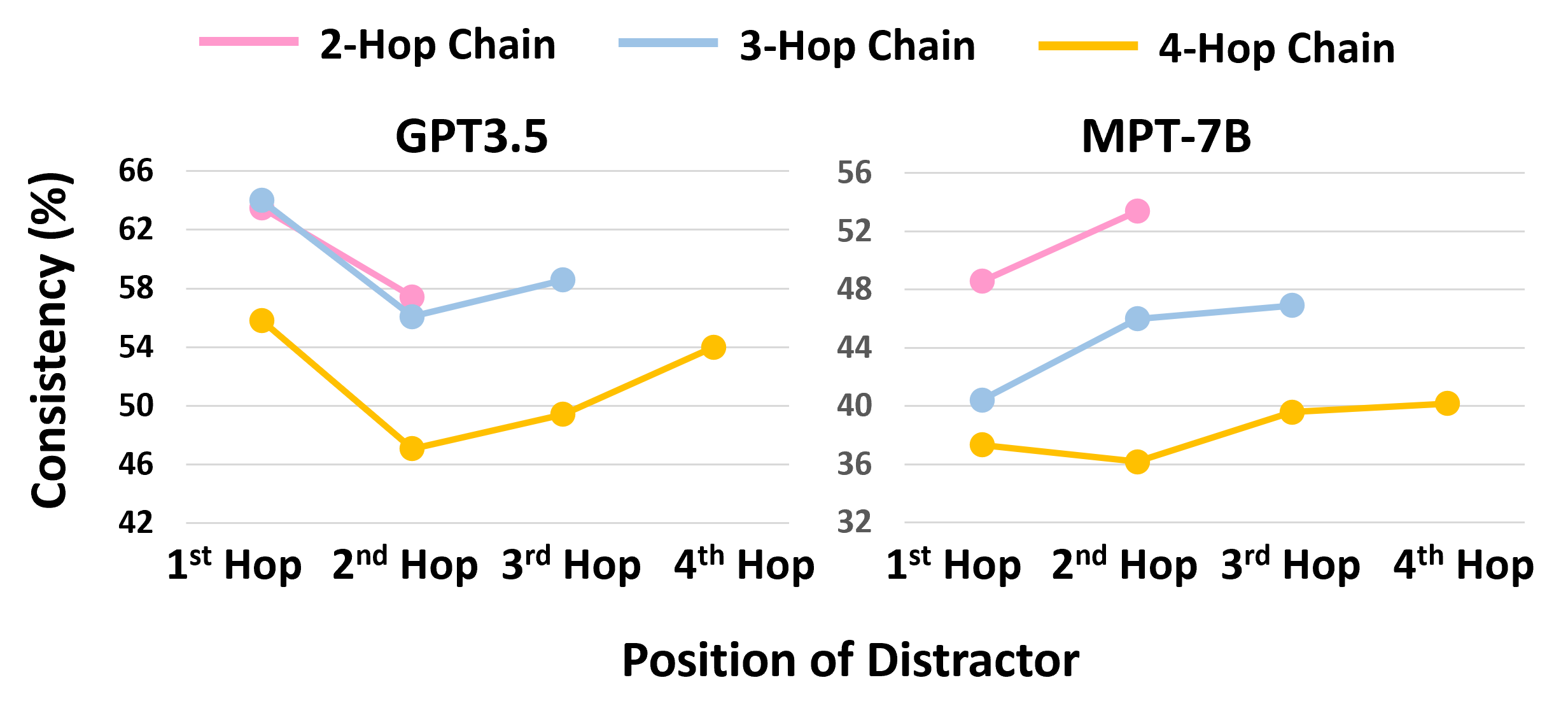}
    \caption{The consistency with respect to which position is the distractor introduced to. MPT-7B's consistency rises gradually as the position to introduce the distractor moves backward, while GPT3.5 shows a high consistency if the distractor is introduced to the first hop (beginning) of the query chain.}
    \label{position_consistency}
\end{figure}

\begin{figure}[!t]
    \centering
    \includegraphics[width=1.0\linewidth]{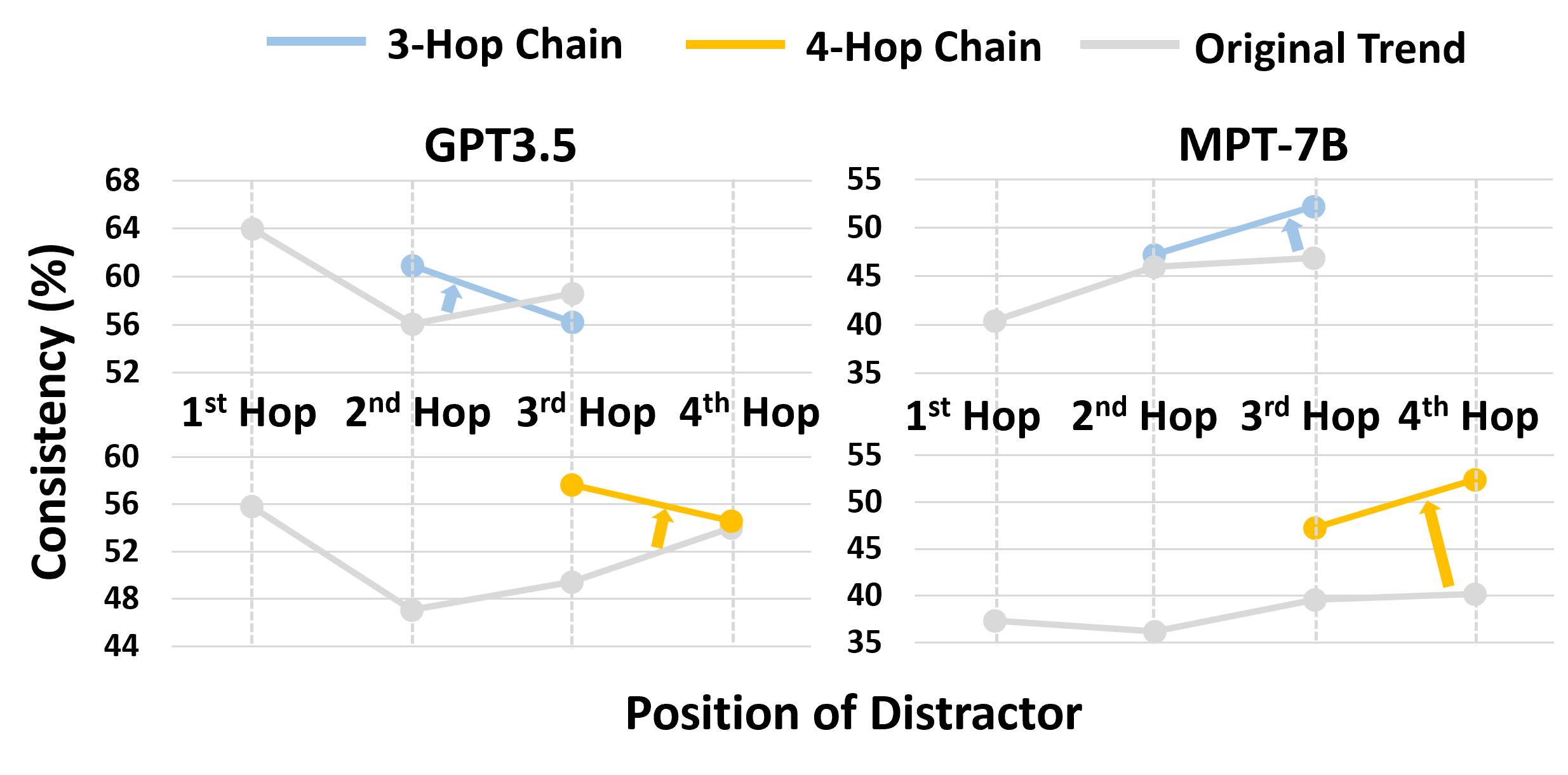}
    \caption{The consistency with respect to querying only the last two hops in the 3-hop and 4-hop chains. While MPT-7B still maintains an upward trend, GPT3.5 exhibits a downward trend similar to that of a 2-hop chain (\textcolor[RGB]{255,153,204}{pink}) in Figure~\ref{position_consistency}.}
    \label{position_support}
\end{figure}

% ==================================== %

\paragraph{Distract Positions: Models resist distractions as interaction evolves, and GPT3.5 also defends against early distractions.} We discover the consistency of both models generally rises as the position where the distractors are introduced moves toward the tail of the data chain. In addition, GPT3.5 is more likely to resist external knowledge if it is introduced in the beginning. To better illustrate the trends, we differentiate three knowledge structures and plot their consistency respectively in Figure~\ref{position_consistency}. For both models, the gradual rising trend could be attributed to their declining attention towards the distractor as the query goes on: as the chatting history accumulates, both models become harder to deviate, thus introducing a distractor that interferes with later hops raises consistency. In addition, the high consistency of GPT3.5 if distracted initially implies it maintains a heightened sensitivity to the veracity of external knowledge, but MPT-7B lacks such a mechanism.

To mitigate the influence of queries themselves, we conduct an ablation study by querying only the last two hops of the 3-hop and 4-hop chains. From the results in Figure~\ref{position_support}, we discover that: i) Overall consistency increases as a shorter data chain is applied. ii) While MPT-7B's consistency still rises, GPT3.5's consistency declines. The higher consistency observed when distractors are introduced to interfere with the first hop provides additional evidence of GPT3.5's heightened vigilance in the beginning towards information that deviates from its PKG. We provide additional detailed results in \cref{Apdx_Distrcat_Positions}.

% ==================================== %

\begin{table}[!t]
\centering
\small
\tabcolsep=0.012\linewidth
\begin{tabular*}{\linewidth}{cccccc}
\toprule
\multirow{2}{*}{\textbf{Metrics}}         & \multirow{2}{*}{\textbf{\makecell{}}} & \multicolumn{2}{c}{\underline{\space\textbf{Conforming Res.}\space}} & \multicolumn{2}{c}{\underline{\space\textbf{Deviated Res.}\space}} \\
                                &  & \texttt{Match}   & \texttt{Shift} & \texttt{Match} & \texttt{Shift} \\
\toprule
\multirow{3}{*}{\textbf{\makecell{Change of\\Confidence\\(\%)}}}
& 2-hop &  \red{-1.55}  &   \red{-2.34}  &   \red{-2.68}  &   \green{+2.42}  \\
& 3-hop &  \red{-1.25}  &   \red{-0.69}  &   \red{-0.50}  &   \green{+5.69}  \\
& 4-hop &  \red{-0.94}  &   \red{-0.21}  &   \green{+0.51}  &   \green{+0.40}  \\
\bottomrule
\end{tabular*}
\caption{The \textit{change} of confidence with respect to distractors of different degrees when the knowledge format becomes lengthier and more detailed. We discover the model's confidence rises mostly in the deviated responses to \textit{Type Shift Distractors}.}
\label{format_degree_analysis}
\end{table}

\begin{table}[!t]
\centering
\small
\tabcolsep=0.0065\linewidth
\begin{tabular*}{\linewidth}{cccccccc}
\toprule
\multirow{2}{*}{\textbf{Metrics}}         & \multirow{2}{*}{\textbf{\makecell{}}} & \multicolumn{3}{c}{\underline{\space\textbf{Conforming Res.}\space}} & \multicolumn{3}{c}{\underline{\space\textbf{Deviated Res.}\space}} \\
&  & \texttt{Obj.} & \texttt{Indir.} & \texttt{Sbj.} & \texttt{Obj.} & \texttt{Indir.} & \texttt{Sbj.} \\
\toprule
\multirow{3}{*}{\textbf{\makecell{Change of\\Confidence\\(\%)}}}
& 2-hop &  \red{-2.21}  &   \red{-1.55}  &   \red{-2.13}  &   \red{-4.52}  &  \green{+3.58}  &   \green{+1.80}  \\
& 3-hop &  \red{-1.24}  &   \red{-0.67}  &   \red{-0.91}  &   \red{-1.93}  &  \red{-0.28}  &   \green{+9.26}  \\
& 4-hop &  \red{-0.66}  &   \red{-1.03}  &   \red{-0.56}  &   \red{-1.80}  &  \red{-0.16}  &   \green{+5.03}  \\
\bottomrule
\end{tabular*}
\caption{The change of confidence with respect to distractors applying different methods when the knowledge format becomes lengthier and more detailed. \textit{Obj.}, \textit{Indir.} and \textit{Sbj.} respectively denotes \textit{Object}, \textit{Indirect}, and \textit{Subject Distractors}. We discover the model's confidence tends to rise on deviated responses for ``weaker'' distract methods.}
\label{format_method_analysis}
\end{table}

% ==================================== %

\paragraph{Distract Formats: Lengthier distract contexts lower the consistency.} We discover that both black-box and open-source models are more susceptible to placing \textit{false} trust in lengthier external knowledge that is seemingly more compelling. This is proved by comparing the consistency of all multi-hop structures between introducing \textit{Single Sentence} and \textit{Paragraph} as external knowledge formats of distractors. The results presented in the fourth row of Table~\ref{results_all_avg}A demonstrate a consistent trend: the consistency decreases for both GPT3.5 and MPT-7B models ($p<0.001$ in both cases, detailed in \cref{Apdx_Distrcat_Formats}) when the context of the distractor is longer and more detailed. 
%We conduct more analysis on the impact of knowledge format together with other distractors' attributes in Section~\ref{Mutual_Impacts}, and provide additional detailed results in Appendices~\ref{Apdx_Distrcat_Formats}.

Why would the model's belief change, even when the core content of the distractor stays the same?
We further investigate the interactions between the format and other distractor attributes. 
We find that \emph{lengthier context raises belief in more severely edited external knowledge}, introduced through \textit{Type Shift Distractors}.
%Taking a step further, we investigate what mainly causes this change in the model's belief \emph{when the core content of the distractor stays the same}.
%We discover that more detailed contexts tend to raise the model's confidence in believing \textit{Type Shift Distractors}, which often lack veracity.
Specifically, we divide the external knowledge based on \textit{Type Match} and \textit{Type Shift}, and examine the resulting \textit{changes} in confidence in both deviated and conforming responses caused by the alteration in the external knowledge format. In Table~\ref{format_degree_analysis}, we observe that MPT-7B's confidence rises mostly in its deviated responses to \textit{Type Shift Distractors}, while its confidence decreases in all conforming responses. This implies that, in general, making the context lengthier lowers the model's confidence in extracting a target entity as the answer in response. However, it also boosts the model's confidence in trusting the \textit{Type Shift Distractors}, which are more severely edited external knowledge.

Besides different degrees, we also discover that \emph{the more detailed contexts tend to raise the model's belief in ``weaker'' distract methods}. Together with the previous finding, it could be concluded that lengthier and more detailed contexts are effective in making the model trust the knowledge that it previously tended not to believe in.
In Table~\ref{format_method_analysis}, we show the resulting change of confidence when the knowledge format becomes \textit{Paragraph} with respect to different distract methods. We discover that MPT-7B's confidence generally rises in its deviated responses when applying \textit{Indirect Distractors} or \textit{Subject Distractors}, which are both ``weaker'' distract methods.

% ==================================== %

\begin{figure}[!t]
    \centering
    \includegraphics[width=\linewidth]{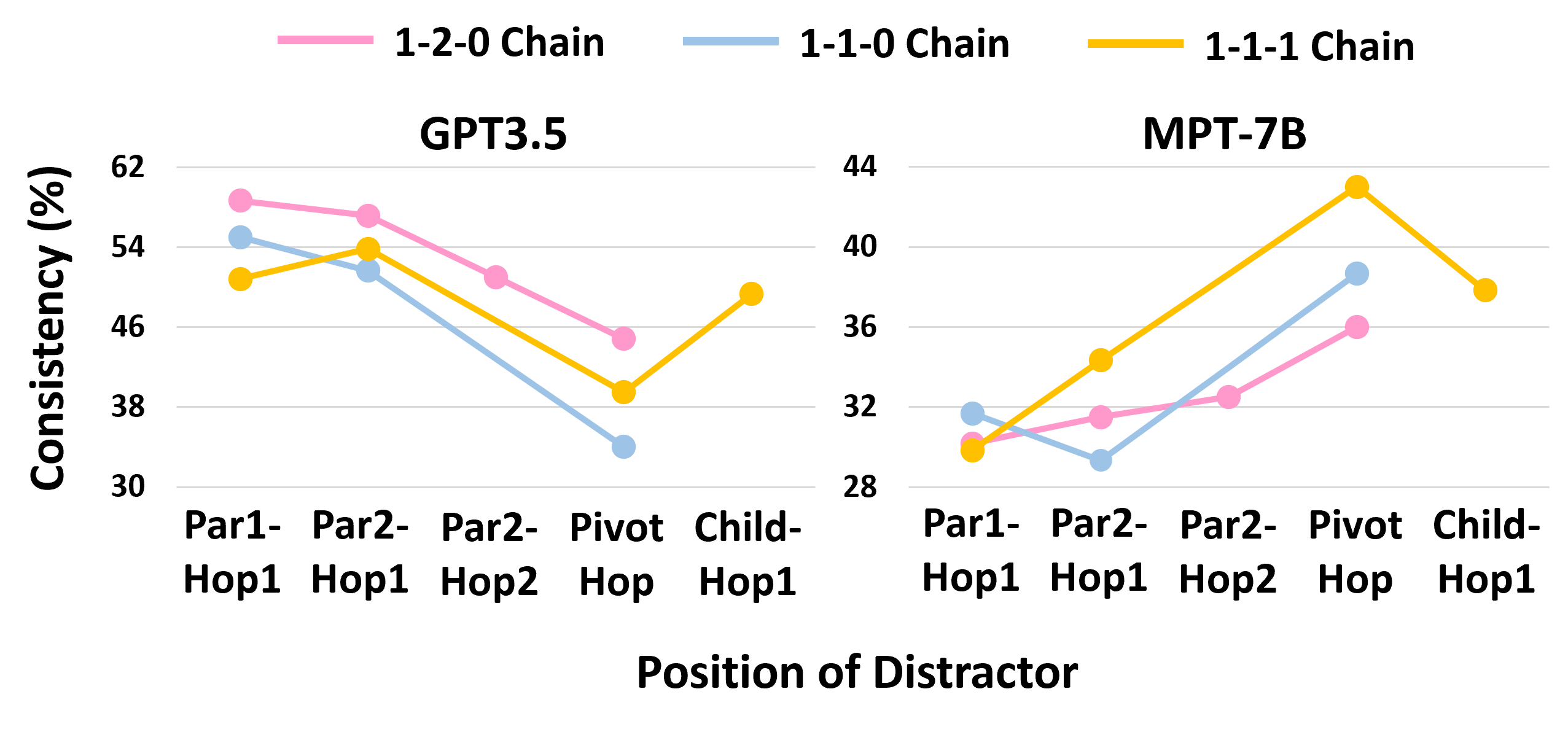}
    \caption{The consistency of GPT3.5 and MPT-7B when applying distractors to different positions in three multi-dependent structures. \textit{Par} denotes the chains extended from the parent nodes of the pivot hop. The results show when the distractor is introduced to interfere with the pivot hop, GPT3.5's consistency is the lowest while MPT-7B's consistency is the highest.}
    \label{structure_position}
\end{figure}

% ==================================== %

\subsection{Results on More PKG Structures}

All previous analyses employ the multi-hop structures within the model's PKG, primarily focusing on one-to-one relations. In this section, we extend the distractor's influence upon multi-dependent structures as introduced in \cref{Setup_Data}, and investigate the impacts brought by the sub-graphs of PKG with various knowledge structures.

\paragraph{Results for distract methods and degrees remain consistent.} Our previous conclusions on distract methods and degrees are further supported by multi-dependent structures. Please refer to the consistent trend in Table~\ref{results_all_avg}B and \cref{Apdx_Multi_Structures} for details.

\paragraph{Pivot hop exhibits unique traits.} The results of introducing distractors on different positions are incomparable to previous conclusions, as the underlying knowledge structure changes. For multi-dependent structures, we discover that GPT3.5 achieves the lowest consistency if distracted on the \textit{pivot} hop, while MPT-7B achieves the highest consistency.

To better observe the trend, we again differentiate three multi-dependent structures, and present our detailed results in Figure~\ref{structure_position}. Both 1-1-0 and 1-1-1 structures do not contain \textit{Par2-Hop2} as the two parent nodes for \textit{pivot} hop are only the ends for two 1-hop queries. Both 1-1-0 and 1-2-0 structures do not contain \textit{Child-Hop1} as the answer nodes for their \textit{pivot} hops are also the ending nodes for the whole chain. As the pivot hop depends on two upstream entities, the resulting trend in Figure~\ref{structure_position} may imply that GPT3.5 is more easily deviated by distractors with auxiliary external information, while vice versa for MPT-7B.

\section{Discussions}

%\sherry{maybe we can merge conclusion and discussion? Or add a starting sentence here to give an overview} \cheng{Help Needed: Now we discuss some especially interesting points in the experiment, for the first and second conclusions, there are no explicit studies on what we have found; while for the third conclusion, I am sure it's commonly seen, but I put here as it's the closest to application. Based on these justifications, how to further introduce some citations / implications?}

Below, we highlight three rather unexpected observations in our study and their implications.

\paragraph{Interference of Indirect Distraction}
While we can expect direct conflicts to cause the model's inconsistency, we are surprised that indirect distractions could also mislead the model (Figure~\ref{method_cases}).
Their success could mainly be attributed to the model's lack of confidence in the original PKG: If the model's confidence about this particularly queried relation in PKG is low, but about \emph{the relation in external knowledge is high}, then the model might deviate even if the external information is not related to what the question asks.
We further identify two reasons why the model tends to believe in external knowledge.
i) \textbf{Veracity}: \textit{Indirect Distractor} could be a fact (as shown in Row A of Figure~\ref{fig3}). Compared to other methods, \emph{the model's ``faith'' in the correctness of external knowledge makes it doubt its original answers}.
ii) \textbf{Matched Details}: The models are more easily distracted by similar details presented in both the query and the external information (e.g., in Figure~\ref{method_cases}, ``1948'' appears in both the distractor and the user's query for MPT-7B). These matching details lead the model to believe that there is a strong correlation between the distractor and the query, thus causing the model to select an entity from external knowledge as the answer.

% Two factors might have contributed to their success: 
% i) \textbf{Lack of confidence in PKG}: If the model's confidence about this particularly queried relation in PKG is low, but about \emph{the relation in external knowledge is high}, then the model might deviate even if the external information is not related to what the question asks. 
% This is more pronounced if the \textit{Indirect Distractor} is actually a fact that the model believes in (unlike the other two distract methods, \textit{Indirect Distrcator} could be fact, as shown in the Row A of Figure~\ref{fig3}).
% In other words, \emph{The model's ``faith'' in the correctness of external knowledge makes it doubt its original answers}.
% ii) \textbf{Matched supporting details}: The models are more easily distracted by similar details presented in both the query and the external information (e.g., ``1948'' appears in both the distractor and the user's query for MPT-7B). These details lead the model to believe that there is a strong correlation between the distractor and the query, thus causing the model to select an entity from the distractor as the answer.

% \sherry{I feel like 2 is covering a lot of points already in 1} \cheng{The first is only about the comparison of confidence: high in external but low in original. But high in external can be attributed to  }
%These details may blind the model, making them ignorant of the fact that even the subject of the question and the distractor are not the same.

The impact of indirect distraction flags additional challenges in misinformation, as most efforts tend to focus on preventing LLMs from trusting wrong statements.
Future studies on effectively removing information snippets that pose unexpected effects will be valuable in the space.

%This observation flags the lack of robustness of LLMs, but it might point to an interesting future venue for studying confidence calibration: 
%How LLM's confidence changes when unrelated factual information

\paragraph{Bias Towards Lengthier Context} 
Both GPT and MPT models demonstrate low consistency when presented with lengthier and more detailed knowledge formats. This implies that the model's judgment is not solely rooted in facts and veracity, but rather resembles human decision-making, influenced by persuasiveness. 
Though lacking decisive evidence, we see that the model is more inclined to accept external knowledge it was previously unaware of or had doubts about, rather than blatantly false information (e.g., ``The Sun rises in the west''). 
This tendency becomes more pronounced when such information is provided within a specific and detailed context with supporting details.
This inherent bias in LLMs directly influences our approach to using them and raises concerns about potential misuse, whether external information is introduced explicitly or implicitly.
Verification methods that compare model behaviors on long prompts vs. short but equally informative prompts (e.g., a high-quality summarization) might be a useful layer for rectifying LLM outputs.

\paragraph{GPT vs. MPT: GPT's Initial Distrust in External Knowledge} We have discovered a different trend between GPT and MPT models with respect to the position to which distractors are introduced. GPT3.5 maintains a high consistency if the distraction is introduced at the beginning of the data chain, and we show in \cref{Apdx_Davinci_Position} that GPT3 also exhibits a similar trend. This vigilance and distrust of external information in the beginning is not exhibited in MPT-7B, and may be attributed to the training process or some protection mechanisms behind the GPT black-box models. 
From our results, we also observe that GPT models are most likely to deviate in the 2$^{nd}$ hop. This implies that though vigilant to the information's veracity initially, GPT's attention will decrease as the reasoning or interaction goes on, which heightens the risk of hallucination.
Unfortunately, we have yet to reveal the root cause of such behavior differences, but we hope to look more into comparing different LLMs.

% ======================================= %

\section {Conclusions}
This paper mainly investigates the impacts of external knowledge on parametric knowledge through systematic experiments. We build a framework that reveals the model's \textit{parametric knowledge graph} and automatically builds the \textit{external knowledge distractors} as the source of interference. We conduct controlled experiments that investigate the impacts of external knowledge's distract degrees, methods, positions, and formats on various parametric knowledge structures including multi-hop and multi-dependent ones. Our results on both GPT3.5 and MPT show that both models tend to provide responses that deviate from their original PKG when the external information poses direct conflicts (\textit{Object Distractors}), gives confounding changes that are not obviously false (\textit{Type Match Distractors}), or provides external knowledge in detailed and lengthier context (\textit{Paragraph Distractors}). In addition, we discover that GPT models are vigilant to external information's veracity in the beginning (\textit{Distractors at 1$^{st}$ Hop}), and that both models are susceptible to even unrelated external knowledge (\textit{Indirect Distractors}). These studies reveal the mechanism of how LLMs handle potential conflicts, and imply the potential risk of hallucination as LLMs integrate external knowledge, even introduced implicitly. We hope our framework can serve as the testbed for more insightful investigations into the active interaction between external and parametric knowledge.

\bibliography{anthology,custom}
\bibliographystyle{acl_natbib}

\clearpage
\appendix

\section*{Appendix}
\label{sec:appendix}

\section{Details on Revealing Model's PKG}
\label{Apdx_PKG_Construct}

We reveal the model's PKGs using natural language templates, as we show in Figure~\ref{example_rules}. During the construction of PKGs, we query the model three times with different temperatures ($T = 0.3, 0.5$, and $0.7$). The prompt for retrieving the answer is shown in Figure~\ref{prompt_answer}. Then, we judge the consistency of the model's responses through the checker we present in Figure~\ref{prompt_consistency}. If we finally get ``N/A'', then we move on to search for other relations complying with the rules that the models are more confident about. Otherwise, we add the relation and the model's consistent answer into the PKG, regarding it as a piece of model's parametric knowledge.

\section{Details on Constructing Distractors}
\label{Apdx_Distractor_Construct}

After the extraction of multiple structures as the original data chain from the model's PKGs, we perform modifications to the data chain to introduce distractors as external knowledge. This process is automated with the help of GPT3.5. Among the four types of distractors, distracting methods and degrees both directly modify the original information. Three distracting methods and two distracting degrees combine into a total of six types of distractors. The prompts applied for constructing these six types of distractors are introduced from Figure~\ref{prompt_object_match} to Figure~\ref{prompt_indirect_shift}.

Upon getting these six types of distractors, the external information we get is in a format of \textit{Single Sentence}. To turn them into external knowledge presented in multiple sentences, we apply the prompt in Figure~\ref{prompt_format} to construct distractors in \textit{Paragraph} format. Distractors introduced to interfere with different positions do not need additional construction.

\section{Details on Experimental Settings}
\label{Apdx_Exp_Setting}
We apply both GPT3.5 and MPT-7B models. For all the experiments, we set \textit{top-p} to 1 and \textit{temperature} to 0.3. The same setting across all experiments guarantees fairness when we are measuring the model's consistency and ensures that the model's confidence is comparable. We set the max sequence length to 512, and for both models, we do not add the frequency or presence penalty.

As introduced in \cref{Apdx_Distractor_Construct}, combining distract methods and degrees, we get six different types of distractors for each hop of the query (each data chain may have multiple hops of the query). We experiment with all these distractors. For the results on distract degrees, we divide the results based on the two different degrees of the distractors. For the results on distract methods, we divide our results based on the three different methods applied in the distractors. For the results on distract positions, we divide the results based on which hop of query in the knowledge structure the distractor is introduced to. For the results on distract knowledge formats, we introduce a \textit{Paragraph} version to all previous distractors and repeat all the experiments for comparison.

\section{Supporting Analysis to Main Results}
For some of the main results, we also perform additional analysis to further support our claims.

% ==================================== %

\begin{table}[!t]
\centering
\small
\tabcolsep=0.05\linewidth
\begin{tabular*}{\linewidth}{cccc}
\toprule
\multirow{2}{*}{\textbf{Metrics}}         & \multirow{2}{*}{\textbf{\makecell{}}} & \multicolumn{2}{c}{\underline{\space\space\space\space\textbf{GPT3.5}\space\space\space\space}} \\
&  & \texttt{Match}  & \texttt{Shift} \\
\toprule
\multirow{3}{*}{\textbf{\makecell{Consistency\\(\%)}}}
& 2-hop  &  59.92  &  $61.00_{\green{\uparrow 1.1}}$  \\
& 3-hop  &  56.78  &  $62.33_{\green{\uparrow 5.6}}$  \\
& 4-hop  &  51.00  &  $52.17_{\green{\uparrow 1.2}}$  \\
\toprule
\multirow{2}{*}{\textbf{Metrics}}         & \multirow{2}{*}{\textbf{\makecell{}}} & \multicolumn{2}{c}{\underline{\space\space\space\space\textbf{MPT-7B}\space\space\space\space}} \\
&  & \texttt{Match}  & \texttt{Shift} \\
\toprule
\multirow{3}{*}{\textbf{\makecell{Consistency\\(\%)}}}
& 2-hop  &  48.25  &  $53.92_{\green{\uparrow 5.7}}$  \\
& 3-hop  &  42.22  &  $46.67_{\green{\uparrow 4.5}}$  \\
& 4-hop  &  37.88  &  $38.75_{\green{\uparrow 0.9}}$  \\
\midrule
\multirow{3}{*}{\textbf{\makecell{Confidence\\(\%)}}}
& 2-hop  &  82.07  &  $78.86_{\red{\downarrow 3.2}}$  \\
& 3-hop  &  80.99  &  $77.78_{\red{\downarrow 3.2}}$  \\
& 4-hop  &  79.71  &  $78.40_{\red{\downarrow 1.3}}$  \\
\bottomrule
\end{tabular*}
\caption{The detailed results for GPT3.5 and MPT-7B when confronting \textit{Type Match} or \textit{Type Shift Distractors} as external interfering knowledge. We differentiate multiple structures instead of performing macro-averaging.}
\label{degree_consistency_detailed}
\end{table}

\begin{table}[!t]
\centering
\small
\tabcolsep=0.05\linewidth
\begin{tabular*}{\linewidth}{cccc}
\toprule
\multirow{2}{*}{\textbf{Metrics}}         & \multirow{2}{*}{\textbf{\makecell{}}} & \multicolumn{2}{c}{\underline{\space\space\space\space\textbf{GPT3.5}\space\space\space\space}} \\
&  & \texttt{Match}  & \texttt{Shift} \\
\toprule
\multirow{3}{*}{\textbf{\makecell{Abstention\\(\%)}}}
& 2-hop  &  37.42  &  $73.72_{\green{\uparrow 36.3}}$  \\
& 3-hop  &  37.53  &  $69.76_{\green{\uparrow 32.2}}$  \\
& 4-hop  &  44.05  &  $61.41_{\green{\uparrow 17.4}}$  \\
\midrule
\multirow{3}{*}{\textbf{\makecell{Variation\\(\%)}}}
& 2-hop  &  62.58  &  $26.28_{\red{\downarrow 36.3}}$  \\
& 3-hop  &  62.47  &  $30.24_{\red{\downarrow 32.2}}$  \\
& 4-hop  &  55.95  &  $38.59_{\red{\downarrow 17.4}}$  \\
\toprule
\multirow{2}{*}{\textbf{Metrics}}         & \multirow{2}{*}{\textbf{\makecell{}}} & \multicolumn{2}{c}{\underline{\space\space\space\space\textbf{MPT-7B}\space\space\space\space}} \\
&  & \texttt{Match}  & \texttt{Shift} \\
\toprule
\multirow{3}{*}{\textbf{\makecell{Abstention\\(\%)}}}
& 2-hop  &  4.35  &  $9.22_{\green{\uparrow 4.9}}$  \\
& 3-hop  &  3.85  &  $6.35_{\green{\uparrow 2.5}}$  \\
& 4-hop  &  8.18  &  $9.93_{\green{\uparrow 1.8}}$  \\
\midrule
\multirow{3}{*}{\textbf{\makecell{Variation\\(\%)}}}
& 2-hop  &  95.65  &  $90.78_{\red{\downarrow 4.9}}$  \\
& 3-hop  &  96.15  &  $93.65_{\red{\downarrow 2.5}}$  \\
& 4-hop  &  91.82  &  $90.07_{\red{\downarrow 1.8}}$  \\
\bottomrule
\end{tabular*}
\caption{The detailed error analysis on inconsistent chains for GPT3.5 and MPT-7B when confronting \textit{Type Match} or \textit{Type Shift Distractors} as external interfering knowledge. We differentiate multiple structures instead of performing macro-averaging.}
\label{degree_error_detailed}
\end{table}

\begin{figure}[!t]
    \centering
    \includegraphics[width=1.0\linewidth]{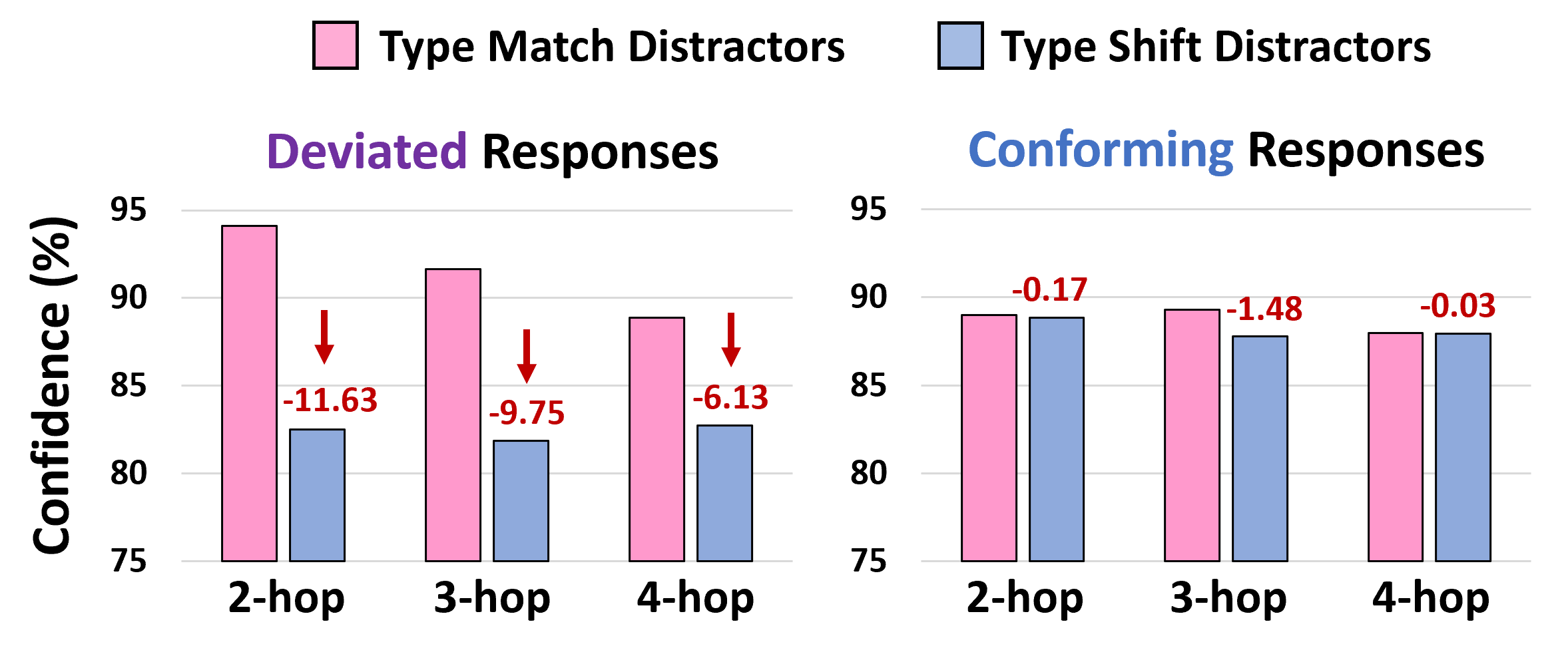}
    \caption{The decrease of confidence brought by changing from \textit{Type Match} to \textit{Type Shift Distractors}. We observe a significant confidence drop for deviated responses when introducing \textit{Type Shift Distractors}, while conforming responses show minor changes.}
    \label{degree_conf}
\end{figure}

% ==================================== %

\subsection{Distract Degrees}
\label{Apdx_Distrcat_Degrees}
The results of the P-value we provide are derived from the T-test between all the consistency values under the interference of \textit{Type Shift Distractors} and \textit{Type Match Distractors}. Each data chain would provide a pair of values for comparison, and there are in total 600 data chains for all 2 / 3 / 4-hop structures.

The results we provide in Table~\ref{results_all_avg} and Table~\ref{degree_error} are macro-average on three structures. We provide more detailed results regarding different multi-hop structures in Table~\ref{degree_consistency_detailed} (Main Metrics) and Table~\ref{degree_error_detailed} (Error Analysis). The consistent trend in every structure provides additional support to our conclusions.

Besides, to further substantiate our claim that the model resists \textit{Type Shift Distractors}, we segment the confidence based on whether the model's response is conforming or deviated. Figure~\ref{degree_conf} displays that the average confidence of generating a deviated response plummets (left chart) when encountering \textit{Type Shift Distractors}, while the confidence of conforming responses shows minor changes. This implies the primary cause of the confidence drop stems from the deviated responses: the model is already hard to be deviated by \textit{Type Shift Distractors}, and for responses that \textit{are} distracted by them, the model's belief in them still remains low.

% ==================================== %

\begin{table}[!t]
\centering
\small
\tabcolsep=0.013\linewidth
\begin{tabular*}{\linewidth}{ccccc}
\toprule
\multirow{2}{*}{\textbf{Metrics}}         & \multirow{2}{*}{\textbf{\makecell{}}} & \multicolumn{3}{c}{\underline{\space\space\space\space\textbf{GPT3.5}\space\space\space\space}} \\
&  & \texttt{Object}  & \texttt{Indirect}  & \texttt{Subject}  \\
\toprule
\multirow{3}{*}{\textbf{\makecell{Consistency\\(\%)}}}
& 2-hop  &  $40.50_{\red{\downarrow 24.8/35.1}}$  &  65.25  &  75.62  \\
& 3-hop  &  $45.75_{\red{\downarrow 17.8/23.7}}$  &  63.50  &  69.42  \\
& 4-hop  &  $42.44_{\red{\downarrow 12.1/15.4}}$  &  54.50  &  57.81  \\
\toprule
\multirow{2}{*}{\textbf{Metrics}}         & \multirow{2}{*}{\textbf{\makecell{}}} & \multicolumn{3}{c}{\underline{\space\space\space\space\textbf{MPT-7B}\space\space\space\space}} \\
&  & \texttt{Object}  & \texttt{Indirect}  & \texttt{Subject}  \\
\toprule
\multirow{3}{*}{\textbf{\makecell{Consistency\\(\%)}}}
& 2-hop  &  $38.38_{\red{\downarrow 11.5/26.6}}$  &  49.88  &  65.00   \\
& 3-hop  &  $35.25_{\red{\downarrow 8.8/18.8}}$  &  44.08  &  54.00   \\
& 4-hop  &  $33.50_{\red{\downarrow 4.4/10.0}}$  &  37.94  &  43.50   \\
\midrule
\multirow{3}{*}{\textbf{\makecell{Confidence\\(\%)}}}
& 2-hop  &  81.87  &  76.67  &  82.92  \\
& 3-hop  &  80.15  &  77.10  &  80.95  \\
& 4-hop  &  79.57  &  77.82  &  79.80  \\
\bottomrule
\end{tabular*}
\caption{The detailed results for GPT3.5 and MPT-7B when confronting \textit{Object}, \textit{Indirect} or \textit{Subject Distractors} as external interfering knowledge. We differentiate multiple structures instead of performing macro-averaging.}
\label{method_consistency_detailed}
\end{table}

\begin{table}[!t]
\centering
\small
\tabcolsep=0.016\linewidth
\begin{tabular*}{\linewidth}{ccccc}
\toprule
\multirow{2}{*}{\textbf{Metrics}}         & \multirow{2}{*}{\textbf{\makecell{}}} & \multicolumn{3}{c}{\underline{\space\space\space\space\textbf{GPT3.5}\space\space\space\space}} \\
&  & \texttt{Object}  & \texttt{Indirect}  & \texttt{Subject}  \\
\toprule
\multirow{3}{*}{\textbf{\makecell{Abstention\\(\%)}}}
& 2-hop  &  $47.06_{\red{\downarrow 16.3/17.0}}$  &  63.31  &  64.10    \\
& 3-hop  &  $48.54_{\red{\downarrow 7.2/7.3}}$  &  55.71  &  55.86    \\
& 4-hop  &  $50.81_{\red{\downarrow 0.4/5.8}}$  &  51.24  &  56.59    \\
\midrule
\multirow{3}{*}{\textbf{\makecell{Variation\\(\%)}}}
& 2-hop  &  $52.94_{\green{\uparrow 16.3/17.0}}$  &  36.69  &  35.90    \\
& 3-hop  &  $51.46_{\green{\uparrow 7.2/7.3}}$  &  44.29  &  44.14    \\
& 4-hop  &  $49.19_{\green{\uparrow 0.4/5.8}}$  &  48.76  &  43.41    \\
\toprule
\multirow{2}{*}{\textbf{Metrics}}         & \multirow{2}{*}{\textbf{\makecell{}}} & \multicolumn{3}{c}{\underline{\space\space\space\space\textbf{MPT-7B}\space\space\space\space}} \\
&  & \texttt{Object}  & \texttt{Indirect}  & \texttt{Subject}  \\
\toprule
\multirow{3}{*}{\textbf{\makecell{Abstention\\(\%)}}}
& 2-hop  &  $8.92_{\green{\uparrow 4.2/3.6}}$  &  4.74  &  5.36    \\
& 3-hop  &  $7.08_{\green{\uparrow 3.5/3.1}}$  &  3.58  &  3.99    \\
& 4-hop  &  $10.24_{\green{\uparrow 2.7/1.0}}$  &  7.55  &  9.29    \\
\midrule
\multirow{3}{*}{\textbf{\makecell{Variation\\(\%)}}}
& 2-hop  &  $91.08_{\red{\downarrow 4.2/3.6}}$  &  95.26  &  94.64    \\
& 3-hop  &  $92.92_{\red{\downarrow 3.5/3.1}}$  &  96.42  &  96.01    \\
& 4-hop  &  $89.76_{\red{\downarrow 3.5/3.1}}$  &  92.45  &  90.71    \\
\bottomrule
\end{tabular*}
\caption{The detailed error analysis on inconsistent chains for GPT3.5 and MPT-7B when confronting \textit{Type Match} or \textit{Type Shift Distractors} as external interfering knowledge. We differentiate multiple structures instead of performing macro-averaging.}
\label{method_error_detailed}
\end{table}

\begin{figure}[!t]
    \centering
    \includegraphics[width=\linewidth]{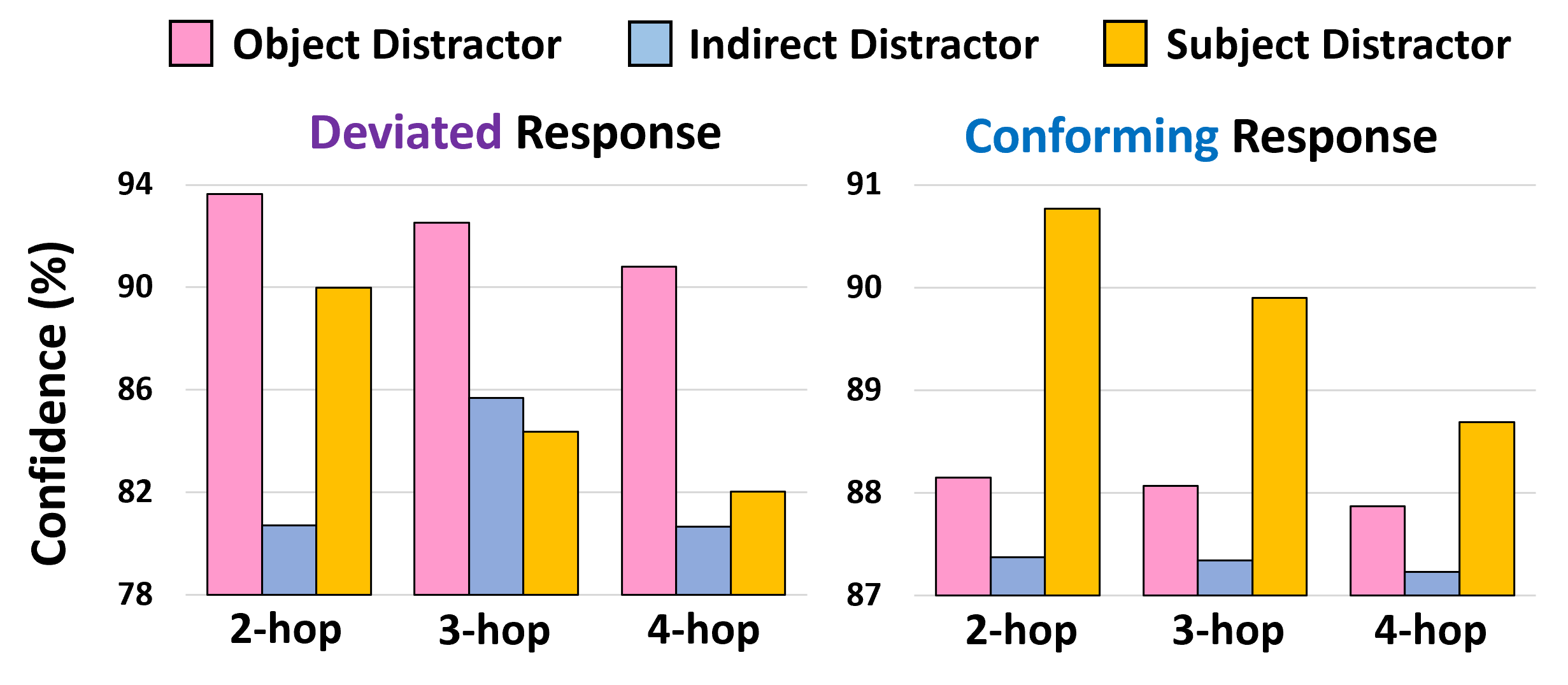}
    \caption{The confidence of MPT-7B's conforming and deviated responses when applying three distract methods. Results show \textit{Object Distractor} induces the highest confidence for deviated responses, while \textit{Subject Distractor} induces the highest confidence for conforming responses.}
    \label{method_conf}
\end{figure}

% ==================================== %

\subsection{Distract Methods}
\label{Apdx_Distrcat_Methods}
We conduct the T-test for all the resulting consistencies between \textit{Object-Indirect Distractors} and \textit{Object-Subject Distractors} for both GPT3.5 and MPT-7B. Similarly, each test comprises 600 pairs of values for comparison.

We provide detailed results regarding the impacts of three distract methods on different multi-hop structures in Table~\ref{method_consistency_detailed} (Main Metrics) and Table~\ref{method_error_detailed} (Error Analysis).

In addition, we analyze MPT-7B's confidence with respect to three distract methods in Figure~\ref{method_conf}. Again, we divide the confidence based on whether the response is conforming or deviated. Our findings show that the \textit{Object Distractor} results in the highest confidence when the response deviates, while the confidence for \textit{Subject Distractor} is highest when the response conforms to the PKG. These results also imply \textit{Object Distractor} is the most powerful to deviate the model's belief, while for the other two ``weaker'' distractors, the model still trusts its original logic pathway in PKG.

% ==================================== %

\begin{table}[!t]
\centering
\small
\tabcolsep=0.010\linewidth
\begin{tabular*}{\linewidth}{cccccc}
\toprule
\multirow{2}{*}{\textbf{Metrics}}         & \multirow{2}{*}{\textbf{\makecell{}}} & \multicolumn{4}{c}{\underline{\space\space\space\space\textbf{GPT3.5}\space\space\space\space}} \\
&  & \texttt{1$^{st}$ Hop}  & \texttt{2$^{nd}$ Hop}  & \texttt{3$^{rd}$ Hop}  & \texttt{4$^{th}$ Hop} \\
\toprule
\multirow{3}{*}{\textbf{\makecell{Consistency\\(\%)}}}
& 2-hop  &  65.50  &  57.42  &  --  &  --  \\
& 3-hop  &  64.00  &  56.08  &  58.58  &  --  \\
& 4-hop  &  55.83  &  47.08  &  49.42  &  54.00  \\
\toprule
\multirow{2}{*}{\textbf{Metrics}}         & \multirow{2}{*}{\textbf{\makecell{}}} & \multicolumn{4}{c}{\underline{\space\space\space\space\textbf{MPT-7B}\space\space\space\space}} \\
&  & \texttt{1$^{st}$ Hop}  & \texttt{2$^{nd}$ Hop}  & \texttt{3$^{rd}$ Hop}  & \texttt{4$^{th}$ Hop} \\
\toprule
\multirow{3}{*}{\textbf{\makecell{Consistency\\(\%)}}}
& 2-hop  &  48.58  &  53.58  &  --  &  --  \\
& 3-hop  &  40.42  &  46.00  &  46.92  &  --  \\
& 4-hop  &  37.33  &  36.17  &  39.58  &  40.17  \\
\midrule
\multirow{3}{*}{\textbf{\makecell{Confidence\\(\%)}}}
& 2-hop  &  78.48  &  82.46  &  --  &  --  \\
& 3-hop  &  78.47  &  80.47  &  79.24  &  --  \\
& 4-hop  &  79.10  &  79.38  &  79.09  &  78.67  \\
\bottomrule
\end{tabular*}
\caption{The detailed results for GPT3.5 and MPT-7B when distractors are introduced in different positions as external interfering knowledge. We differentiate multiple structures instead of performing macro-averaging.}
\label{position_consistency_detailed}
\end{table}

% ==================================== %

\subsection{Distract Positions}
\label{Apdx_Distrcat_Positions}
We provide detailed results on GPT3.5 and MPT-7B's consistency and confidence with respect to different positions where the distractor is introduced in Table~\ref{position_consistency_detailed}. We have plotted the trend of consistency in our main results in Figure~\ref{position_consistency}.

% ==================================== %

\begin{table}[!t]
\centering
\small
\tabcolsep=0.025\linewidth
\begin{tabular*}{\linewidth}{cccc}
\toprule
\multirow{2}{*}{\textbf{Metrics}}         & \multirow{2}{*}{\textbf{\makecell{}}} & \multicolumn{2}{c}{\underline{\space\space\space\space\textbf{GPT3.5}\space\space\space\space}} \\
&  & \texttt{Single Sentence}  & \texttt{Paragraph} \\
\toprule
\multirow{3}{*}{\textbf{\makecell{Consistency\\(\%)}}}
& 2-hop  &  60.46  &  $57.83_{\red{\downarrow 2.6}}$  \\
& 3-hop  &  59.56  &  $56.53_{\red{\downarrow 3.0}}$  \\
& 4-hop  &  51.59  &  $48.75_{\red{\downarrow 2.9}}$  \\
\toprule
\multirow{2}{*}{\textbf{Metrics}}         & \multirow{2}{*}{\textbf{\makecell{}}} & \multicolumn{2}{c}{\underline{\space\space\space\space\textbf{MPT-7B}\space\space\space\space}} \\
&  & \texttt{Single Sentence}  & \texttt{Paragraph} \\
\toprule
\multirow{3}{*}{\textbf{\makecell{Consistency\\(\%)}}}
& 2-hop  &  51.09  &  $48.38_{\red{\downarrow 2.7}}$  \\
& 3-hop  &  44.11  &  $42.34_{\red{\downarrow 1.8}}$  \\
& 4-hop  &  38.32  &  $35.46_{\red{\downarrow 2.9}}$  \\
\midrule
\multirow{3}{*}{\textbf{\makecell{Confidence\\(\%)}}}
& 2-hop  &  80.48  &  $78.28_{\red{\downarrow 2.2}}$  \\
& 3-hop  &  79.40  &  $78.45_{\red{\downarrow 1.0}}$  \\
& 4-hop  &  79.06  &  $78.36_{\red{\downarrow 0.7}}$  \\
\bottomrule
\end{tabular*}
\caption{The detailed results for GPT3.5 and MPT-7B when confronting \textit{Single Sentence} or \textit{Paragraph} as the format of external interfering knowledge. We differentiate multiple structures instead of performing macro-averaging.}
\label{format_consistency_detailed}
\end{table}

% ==================================== %

\subsection{Distract Formats}
\label{Apdx_Distrcat_Formats}
We conduct the T-test for all the resulting consistencies between \textit{Single Sentence} and \textit{Paragraph} as knowledge format for both GPT3.5 and MPT-7B. Similarly, each test comprises 600 pairs of values for comparison.

We provide detailed results on GPT3.5 and MPT-7B's consistency and confidence with respect to the \textit{Single Sentence} or \textit{Paragraph} as the format of external knowledge in Table~\ref{format_consistency_detailed}. We show from the detailed results that every structure's trend is consistent with our main conclusion.

% ==================================== %

\begin{table}[!t]
\centering
\small
\tabcolsep=0.05\linewidth
\begin{tabular*}{\linewidth}{cccc}
\toprule
\multirow{2}{*}{\textbf{Metrics}}         & \multirow{2}{*}{\textbf{\makecell{}}} & \multicolumn{2}{c}{\underline{\space\space\space\space\textbf{GPT3.5}\space\space\space\space}} \\
&  & \texttt{Match}  & \texttt{Shift} \\
\toprule
\multirow{3}{*}{\textbf{\makecell{Consistency\\(\%)}}}
& 1-1-0  &  46.33  &  $47.33_{\green{\uparrow 1.1}}$  \\
& 1-1-1  &  47.75  &  $49.00_{\green{\uparrow 1.3}}$  \\
& 1-2-0  &  51.58  &  $54.25_{\green{\uparrow 2.7}}$  \\
\toprule
\multirow{2}{*}{\textbf{Metrics}}         & \multirow{2}{*}{\textbf{\makecell{}}} & \multicolumn{2}{c}{\underline{\space\space\space\space\textbf{MPT-7B}\space\space\space\space}} \\
&  & \texttt{Match}  & \texttt{Shift} \\
\toprule
\multirow{3}{*}{\textbf{\makecell{Consistency\\(\%)}}}
& 1-1-0  &  32.00  &  $34.44_{\green{\uparrow 2.4}}$  \\
& 1-1-1  &  42.22  &  $46.67_{\green{\uparrow 4.5}}$  \\
& 1-2-0  &  32.00  &  $33.08_{\green{\uparrow 1.1}}$  \\
\midrule
\multirow{3}{*}{\textbf{\makecell{Confidence\\(\%)}}}
& 1-1-0  &  76.88  &  $74.96_{\red{\downarrow 1.9}}$  \\
& 1-1-1  &  79.35  &  $77.82_{\red{\downarrow 1.5}}$  \\
& 1-2-0  &  79.57  &  $77.65_{\red{\downarrow 1.9}}$  \\
\bottomrule
\end{tabular*}
\caption{The detailed results for GPT3.5 and MPT-7B when multi-dependent structures (1-1-0, 1-1-1, and 1-2-0) confronts \textit{Type Match} or \textit{Type Shift Distractors} as external interfering knowledge. We differentiate the three structures instead of performing macro-averaging.}
\label{structure_degree_detailed}
\end{table}

\begin{table}[!t]
\centering
\small
\tabcolsep=0.015\linewidth
\begin{tabular*}{\linewidth}{ccccc}
\toprule
\multirow{2}{*}{\textbf{Metrics}}         & \multirow{2}{*}{\textbf{\makecell{}}} & \multicolumn{3}{c}{\underline{\space\space\space\space\textbf{GPT3.5}\space\space\space\space}} \\
&  & \texttt{Object}  & \texttt{Indirect}  & \texttt{Subject}  \\
\toprule
\multirow{3}{*}{\textbf{\makecell{Consistency\\(\%)}}}
& 1-1-0  &  $32.83_{\red{\downarrow 11.7/30.5}}$  &  44.50  &  63.33  \\
& 1-1-1  &  $32.25_{\red{\downarrow 17.1/28.3}}$  &  49.38  &  60.50  \\
& 1-2-0  &  $41.12_{\red{\downarrow 13.8/21.6}}$  &  54.87  &  62.75  \\
\toprule
\multirow{2}{*}{\textbf{Metrics}}         & \multirow{2}{*}{\textbf{\makecell{}}} & \multicolumn{3}{c}{\underline{\space\space\space\space\textbf{MPT-7B}\space\space\space\space}} \\
&  & \texttt{Object}  & \texttt{Indirect}  & \texttt{Subject}  \\
\toprule
\multirow{3}{*}{\textbf{\makecell{Consistency\\(\%)}}}
& 1-1-0  &  $20.67_{\red{\downarrow 14.3/25.3}}$  &  34.00  &  45.00   \\
& 1-1-1  &  $29.38_{\red{\downarrow 5.2/15.4}}$  &  34.62  &  44.75   \\
& 1-2-0  &  $24.12_{\red{\downarrow 7.1/18.1}}$  &  31.25  &  42.25   \\
\midrule
\multirow{3}{*}{\textbf{\makecell{Confidence\\(\%)}}}
& 1-1-0  &  75.27  &  74.50  &  77.98  \\
& 1-1-1  &  78.81  &  77.63  &  79.33  \\
& 1-2-0  &  79.01  &  77.41  &  79.42  \\
\bottomrule
\end{tabular*}
\caption{The detailed results for GPT3.5 and MPT-7B when multi-dependent structures (1-1-0, 1-1-1, and 1-2-0) confronts \textit{Object}, \textit{Indirect} or \textit{Subject Distractors} as external interfering knowledge. We differentiate the three structures instead of performing macro-averaging.}
\label{structure_method_detailed}
\end{table}

\begin{table}[!t]
\centering
\small
\tabcolsep=0.0155\linewidth
\begin{tabular*}{\linewidth}{ccccccc}
\toprule
\multirow{2}{*}{\textbf{Metrics}}         & \multirow{2}{*}{\textbf{\makecell{}}} & \multicolumn{5}{c}{\underline{\space\space\space\space\textbf{GPT3.5}\space\space\space\space}} \\
&  & \texttt{\makecell{Par1\\Hop1}}  & \texttt{\makecell{Par2\\Hop1}} & \texttt{\makecell{Par2\\Hop2}}  & \texttt{\makecell{Pivot\\Hop}}  & \texttt{\makecell{Child\\Hop1}} \\
\toprule
\multirow{3}{*}{\textbf{\makecell{Consistency\\(\%)}}}
& 2-hop  &  55.00  &  51.67  &  --  &  34.00  &  --  \\
& 3-hop  &  50.83  &  53.83  &  39.50  &  --  &  49.33  \\
& 4-hop  &  58.67  &  57.17  &  51.00  &  44.83  &  --  \\
\toprule
\multirow{2}{*}{\textbf{Metrics}}         & \multirow{2}{*}{\textbf{\makecell{}}} & \multicolumn{5}{c}{\underline{\space\space\space\space\textbf{MPT-7B}\space\space\space\space}} \\
&  & \texttt{\makecell{Par1\\Hop1}}  & \texttt{\makecell{Par2\\Hop1}} & \texttt{\makecell{Par2\\Hop2}}  & \texttt{\makecell{Pivot\\Hop}}  & \texttt{\makecell{Child\\Hop1}} \\
\toprule
\multirow{3}{*}{\textbf{\makecell{Consistency\\(\%)}}}
& 2-hop  &  31.67  &  29.33  &  --  &  38.67  &  --  \\
& 3-hop  &  29.83  &  34.33  &  --  &  43.00  &  37.83  \\
& 4-hop  &  30.17  &  31.50  &  32.50  &  36.00  &  --  \\
\midrule
\multirow{3}{*}{\textbf{\makecell{Confidence\\(\%)}}}
& 2-hop  &  74.48  &  73.72  &  --  &  79.52  &  --  \\
& 3-hop  &  78.11  &  76.24  &  --  &  81.32  &  78.67  \\
& 4-hop  &  78.61  &  77.56  &  78.52  &  79.76  &  --  \\
\bottomrule
\end{tabular*}
\caption{The detailed results for GPT3.5 and MPT-7B when distractors are introduced in different positions of multi-dependent structures (1-1-0, 1-1-1, and 1-2-0) as external interfering knowledge. \textit{Par.} denotes the chains extended from the parent nodes of the \textit{pivot} query. We differentiate the three structures instead of performing macro-averaging.}
\label{structure_position_detailed}
\end{table}

% ==================================== %

\subsection{Multi-Dependent Structures}
\label{Apdx_Multi_Structures}
To establish the overarching applicability of our prior conclusions, we undertake a parallel analysis with distractors of different methods and degrees to multi-dependent structures in PKG. The experimental settings and methods are kept the same as those for multi-hop structures. As delineated in Table~\ref{structure_degree_detailed}, for all three multi-dependent structures, our findings reveal that the model's consistency is higher in response to \textit{Type Shift Distractors}, though the model's overall confidence lowers. Furthermore, Table~\ref{structure_method_detailed} showcases that the \textit{Object Distractors} remain the prime catalyst for the model's deviation. Notably, these insights are consistent with the outcomes obtained from our investigations into multi-hop structures.

Furthermore, we provide detailed results on GPT3.5 and MPT-7B's consistency and confidence with respect to different positions where the distractor is introduced in Table~\ref{structure_position_detailed}. We have plotted the trend of consistency in our main results in Figure~\ref{structure_position}.

\section{Additional Results from GPT3}
\label{Apdx_Exp_Davinci}
To further support our discoveries, we perform additional experiments on GPT3 (Text-Davinci-003). Though GPT3 is not designed as a conversational model, its results can still reflect and bolster some of the trends that we have discovered. We perform experiments on 2 / 3 / 4-hop data chains, with 100 raw chains from GPT3's PKG for each type. We keep the rules we applied for constructing the PKG the same, and we keep all the other experimental setups the same as we introduced in \cref{Apdx_Exp_Setting}.

% ==================================

\begin{table}[!t]
\centering
\small
\tabcolsep=0.05\linewidth
\begin{tabular*}{\linewidth}{cccc}
\toprule
\multirow{2}{*}{\textbf{Metrics}}         & \multirow{2}{*}{\textbf{\makecell{}}} & \multicolumn{2}{c}{\underline{\space\space\space\space\textbf{GPT3}\space\space\space\space}} \\
&  & \texttt{Match}  & \texttt{Shift} \\
\toprule
\multirow{3}{*}{\textbf{\makecell{Consistency\\(\%)}}}
& 2-hop  &  52.67  &  $68.00_{\green{\uparrow 15.3}}$  \\
& 3-hop  &  43.56  &  $52.78_{\green{\uparrow 9.2}}$  \\
& 4-hop  &  46.58  &  $52.92_{\green{\uparrow 6.3}}$  \\
\midrule
\multirow{3}{*}{\textbf{\makecell{Confidence\\(\%)}}}
& 2-hop  &  69.87  &  $66.98_{\red{\downarrow 2.9}}$  \\
& 3-hop  &  68.41  &  $66.05_{\red{\downarrow 2.4}}$  \\
& 4-hop  &  70.04  &  $68.10_{\red{\downarrow 1.9}}$  \\
\bottomrule
\end{tabular*}
\caption{The consistency and confidence of GPT3 when confronting \textit{Type Match} or \textit{Type Shift Distractors} as external interfering knowledge. The conclusion on distract degrees is consistent and even more pronounced for GPT3.}
\label{degree_consistency_davinci}
\end{table}

\begin{figure}[!t]
    \centering
    \includegraphics[width=1.0\linewidth]{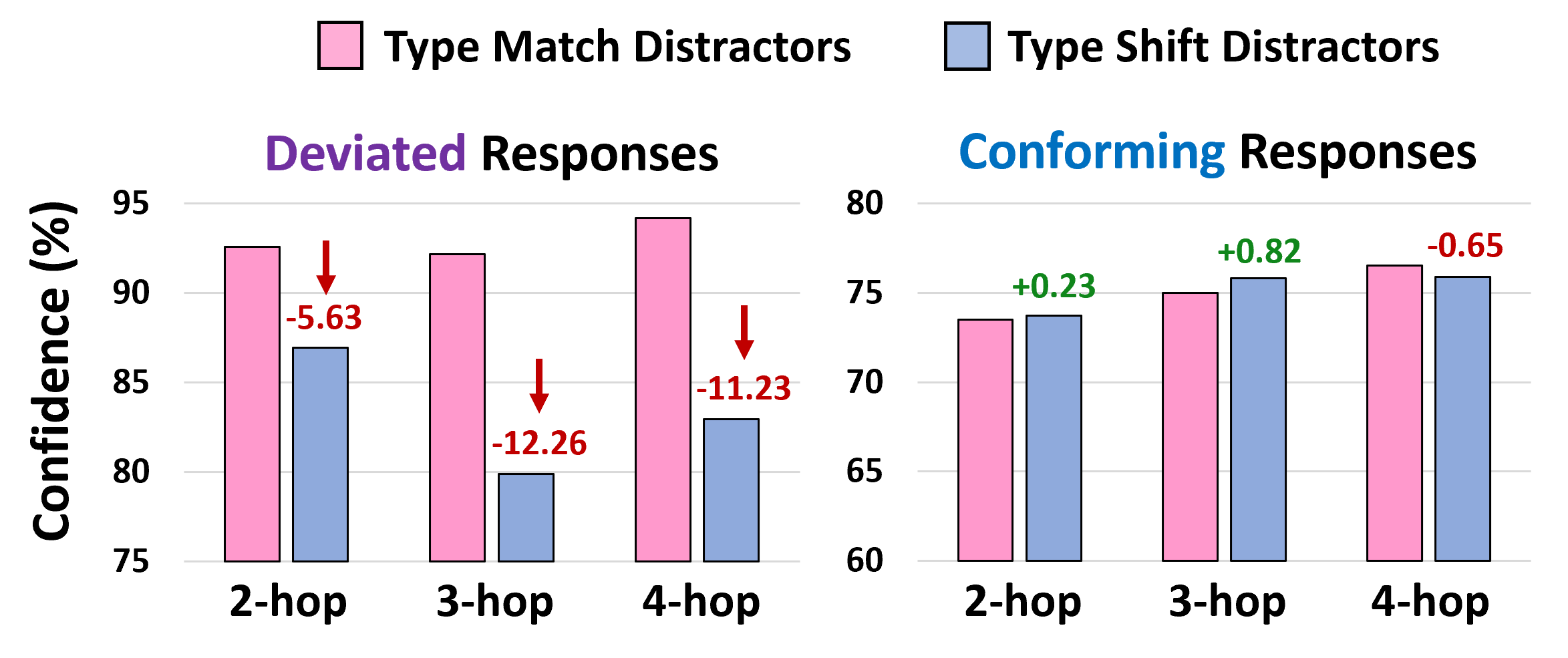}
    \caption{The decrease of confidence in GPT3 brought by changing from \textit{Type Match} to \textit{Type Shift Distractors}. The confidence drop can be mainly attributed to the deviated response, which implies GPT3 also shows resistance to \textit{Type Shift Distractors}.}
    \label{degree_conf_davinci}
\end{figure}

% ==================================

\subsection{Distract Degrees}
In Table~\ref{degree_consistency_davinci}, we observe the same trend in GPT3 that the model resists \textit{Type Shift Distractors} as external knowledge, and the overall confidence in responses is lowered. By further dividing the responses into conforming and deviated ones, we show in Figure~\ref{degree_conf_davinci} that, similarly, the drop in confidence can be mainly attributed to the deviated responses. All these results further bolster the claim that the model put less faith in more severely edited knowledge represented by \textit{Type Shift Distractors}.

% ==================================

\begin{table}[!t]
\centering
\small
\tabcolsep=0.013\linewidth
\begin{tabular*}{\linewidth}{ccccc}
\toprule
\multirow{2}{*}{\textbf{Metrics}}         & \multirow{2}{*}{\textbf{\makecell{}}} & \multicolumn{3}{c}{\underline{\space\space\space\space\textbf{GPT3}\space\space\space\space}} \\
&  & \texttt{Object}  & \texttt{Indirect} & \texttt{Subject} \\
\toprule
\multirow{3}{*}{\textbf{\makecell{Consistency\\(\%)}}}
& 2-hop  &  $40.75_{\red{\downarrow 24.5/34.3}}$  &  65.25  &  75.00   \\
& 3-hop  &  $34.00_{\red{\downarrow 18.5/24.0}}$  &  52.50  &  58.00   \\
& 4-hop  &  $39.00_{\red{\downarrow 14.1/18.1}}$  &  53.12  &  57.12   \\
\midrule
\multirow{3}{*}{\textbf{\makecell{Confidence\\(\%)}}}
& 2-hop  &  71.72  &  61.34  &  72.33   \\
& 3-hop  &  69.79  &  63.30  &  68.62   \\
& 4-hop  &  71.29  &  65.27  &  70.74   \\
\bottomrule
\end{tabular*}
\caption{The consistency of GPT3 when confronting distractors that apply different distract methods. Under the interference of \textit{Object Distractors}, GPT3 shows the lowest consistency. This result still remains consistent with previous conclusions.}
\label{method_consistency_davinci}
\end{table}

% ==================================

\subsection{Distract Methods}
In addition to distractors of different degrees, we also investigate GPT3's consistency towards distractors that apply different methods. In Table~\ref{method_consistency_davinci}, we observe that \textit{Object Distractors} still result in the lowest consistency. This trend also remains the same as what we have shown previously, indicating that GPT3 is also susceptible to \textit{Object Distractors} the most, while the other two ``weaker'' distractors also bring certain impacts.

% ==================================

\begin{figure}[!t]
    \centering
    \includegraphics[width=0.85\linewidth]{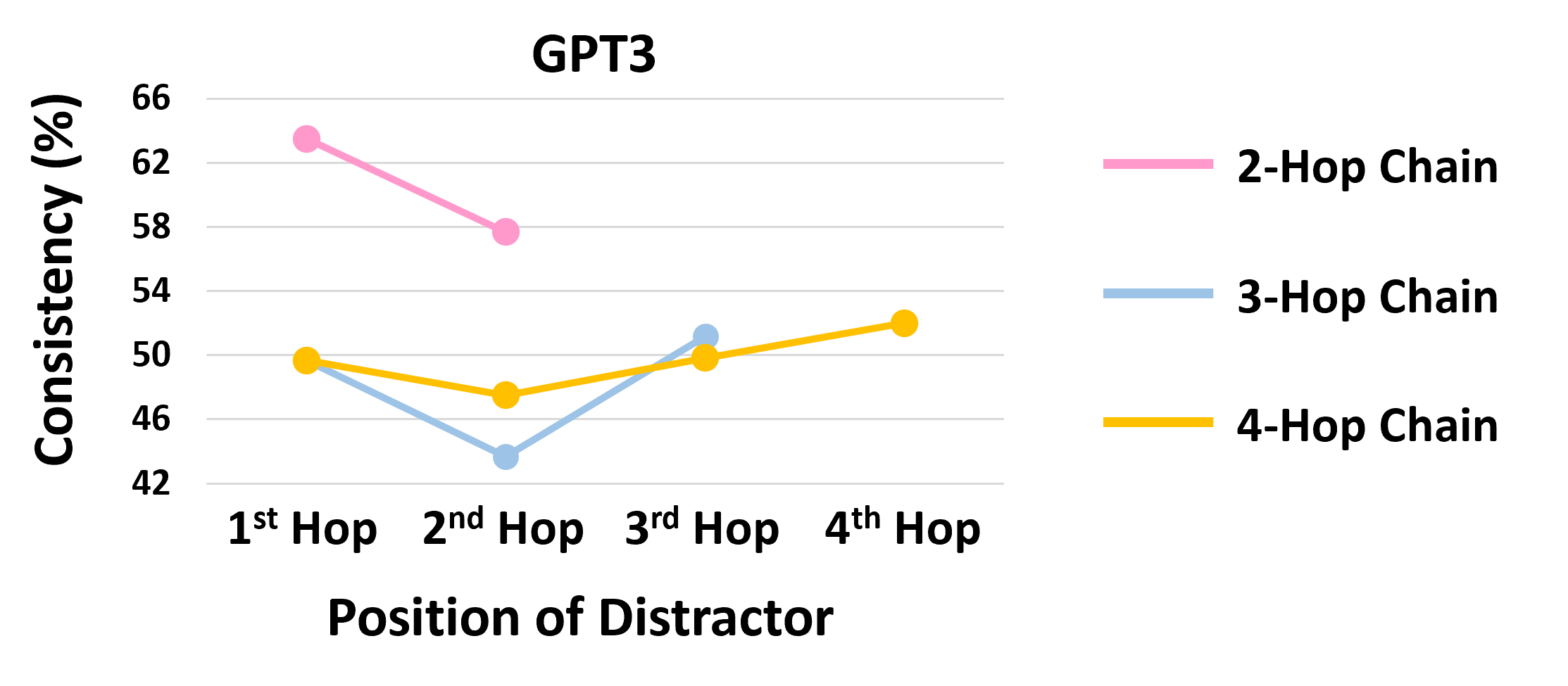}
    \caption{The consistency of GPT3 when the distractor is introduced to interfere with different positions in the data chain. GPT3 exhibits a similar trend as GPT3.5 in Figure~\ref{position_consistency}.}
    \label{position_consistency_davinci}
\end{figure}

% ==================================

\subsection{Distract Position}
\label{Apdx_Davinci_Position}
The pattern for distract positions is different for GPT3.5 and MPT-7B, as we have shown earlier in Figure~\ref{position_consistency}. In Figure~\ref{position_consistency_davinci}, we demonstrate that GPT3's trend is more similar to GPT3.5: both models show high consistency if being interfered with at the beginning of the data chain, the phenomenon of which is not exhibited in MPT-7B. Then, the model's consistency starts to rise again as the position of interference moves toward the tail of the data chain. GPT3's results further support the \textit{GPT family}'s initial sensitivity towards information's veracity.

% ==================================

\begin{table}[!t]
\centering
\small
\tabcolsep=0.025\linewidth
\begin{tabular*}{\linewidth}{cccc}
\toprule
\multirow{2}{*}{\textbf{Metrics}}         & \multirow{2}{*}{\textbf{\makecell{}}} & \multicolumn{2}{c}{\underline{\space\space\space\space\textbf{GPT3}\space\space\space\space}} \\
&  & \texttt{Single Sentence}  & \texttt{Paragraph} \\
\toprule
\multirow{3}{*}{\textbf{\makecell{Consistency\\(\%)}}}
& 2-hop  &  60.34  &  $47.30_{\red{\downarrow 13.0}}$  \\
& 3-hop  &  48.17  &  $40.00_{\red{\downarrow 8.2}}$  \\
& 4-hop  &  49.75  &  $44.00_{\red{\downarrow 5.6}}$  \\
\bottomrule
\end{tabular*}
\caption{The comparison of GPT3's consistency when presented with distractor of \textit{Single Sentence} format versus \textit{Paragraph} format. GPT3's consistency also lowers when the context becomes lengthier.}
\label{format_consistency_davinci}
\end{table}

% ==================================

\subsection{Distract Formats}
We extend the context format of all the distractors into \textit{Paragraph} in the same way we do for GPT3.5 and MPT-7B. We present the results of the comparison in Table~\ref{format_consistency_davinci}. GPT3's consistency lowers in a more pronounced way than GPT3.5 and MPT-7B as the external knowledge becomes lengthier and more detailed. This further supports our previous conclusions and also implies that GPT3 is more biased to trust the detailed but potentially false external knowledge.

% ===============PKG Construction===============

\begin{figure}[!t]
    \centering
    \includegraphics[width=\linewidth]{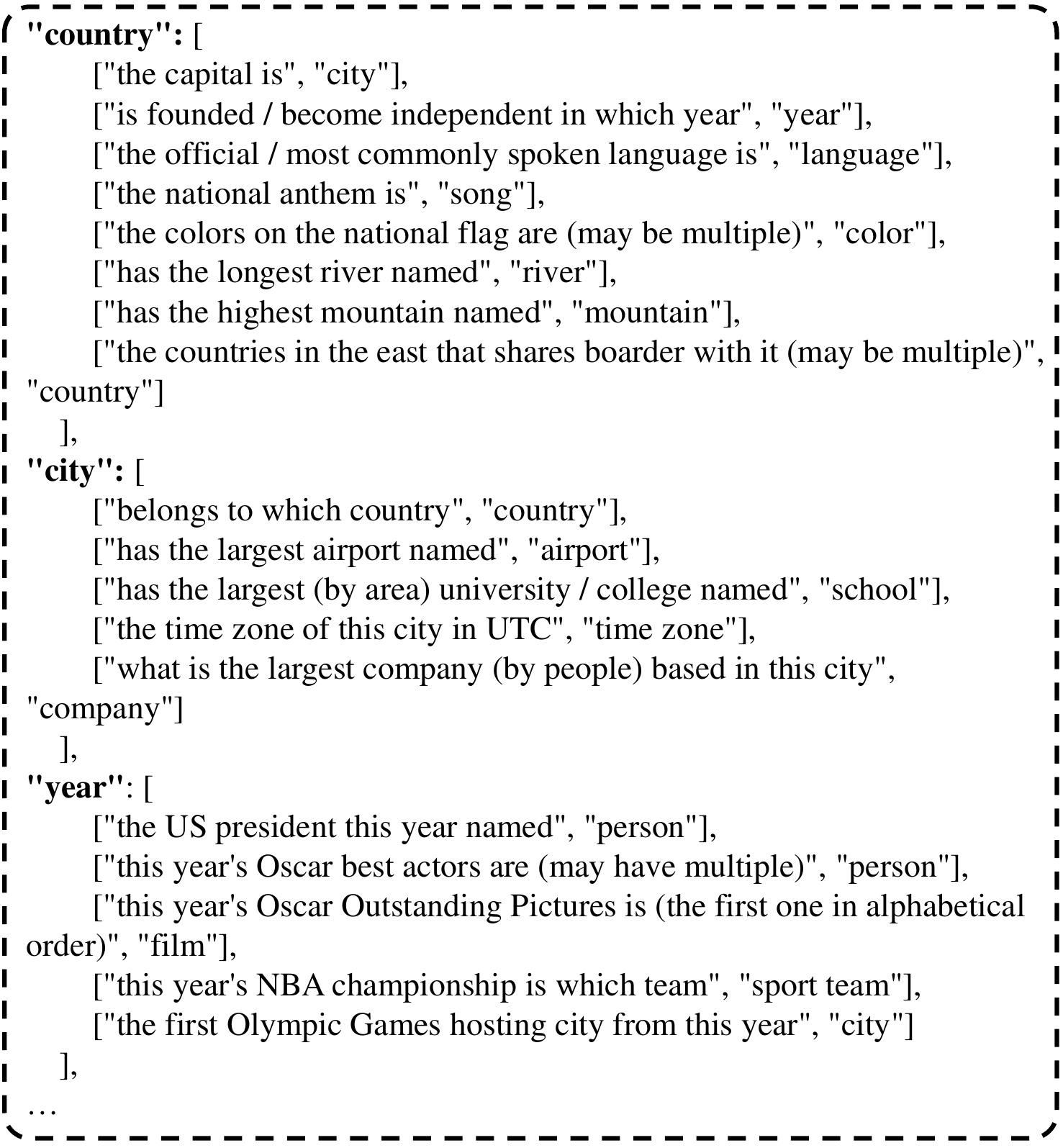}
    \caption{The example rules we apply in building the PKG.}
    \label{example_rules}
\end{figure}

\begin{figure}[!t]
    \centering
    \includegraphics[width=\linewidth]{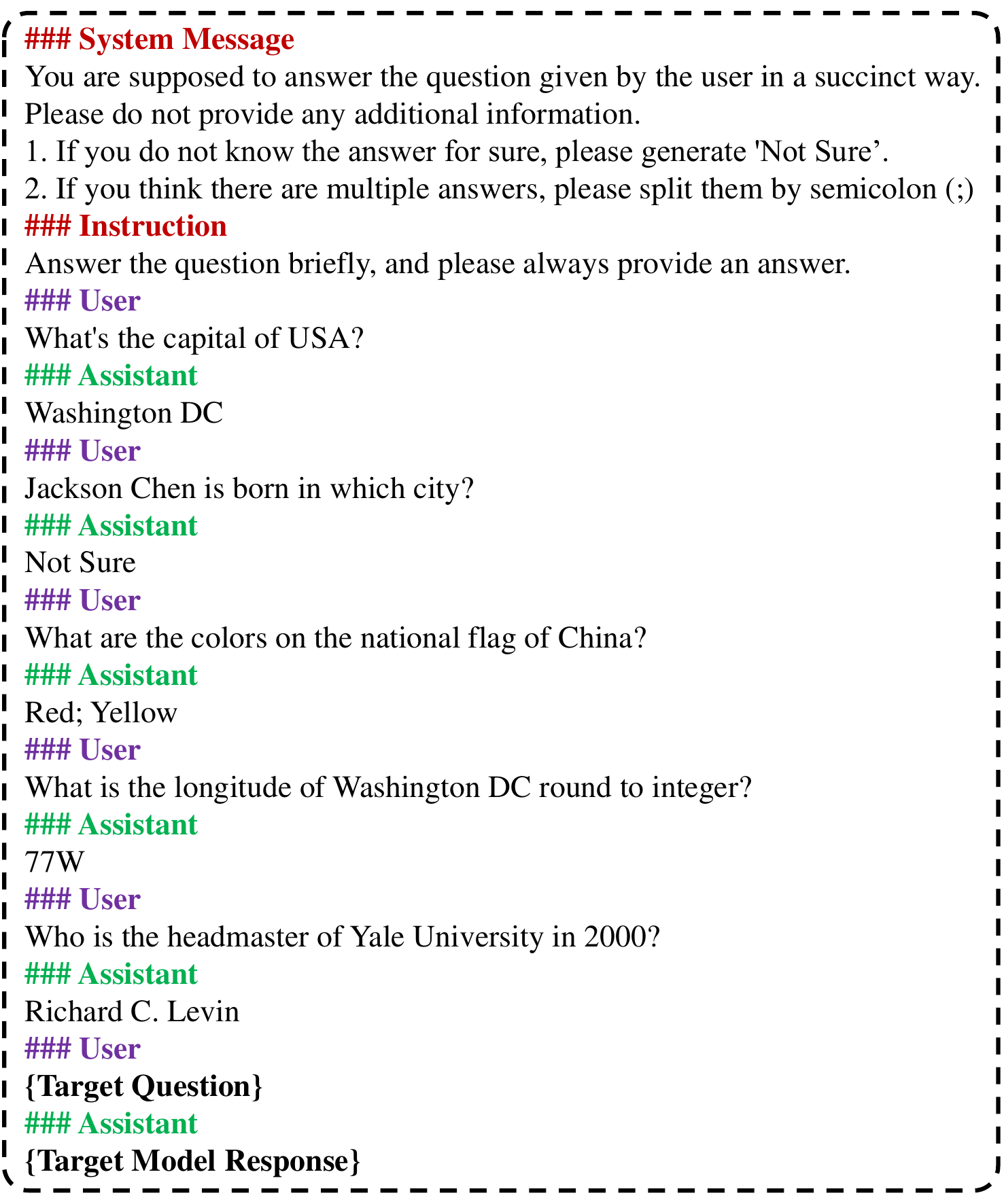}
    \caption{The prompt for retrieving the model's answer when building the PKG.}
    \label{prompt_answer}
\end{figure}

\begin{figure}[!t]
    \centering
    \includegraphics[width=\linewidth]{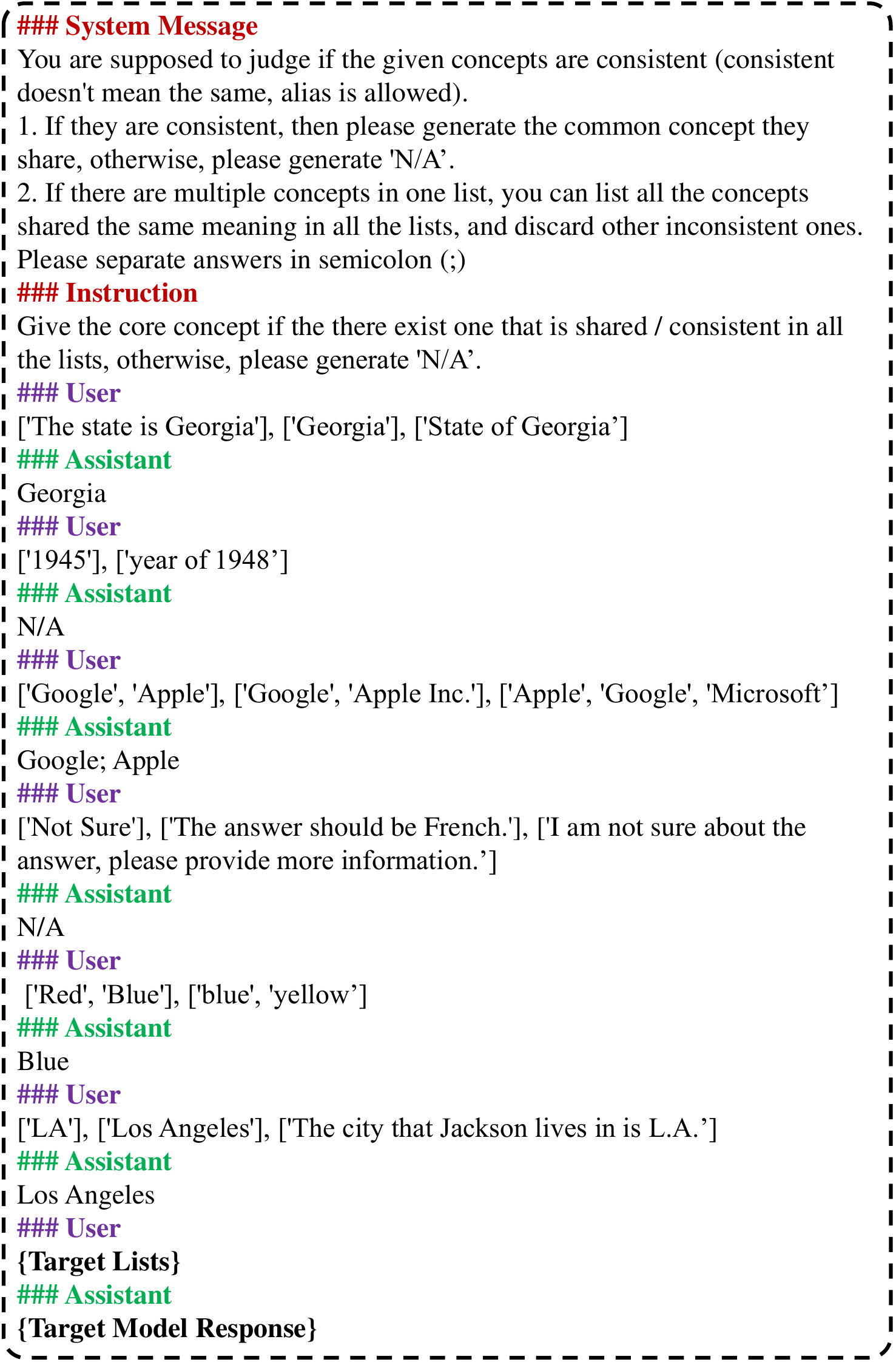}
    \caption{The prompt for judging the consistency of the model's answers when building the PKG.}
    \label{prompt_consistency}
\end{figure}

% ==================================

% ===============External Distractors===============

\begin{figure}[!t]
    \centering
    \includegraphics[width=\linewidth]{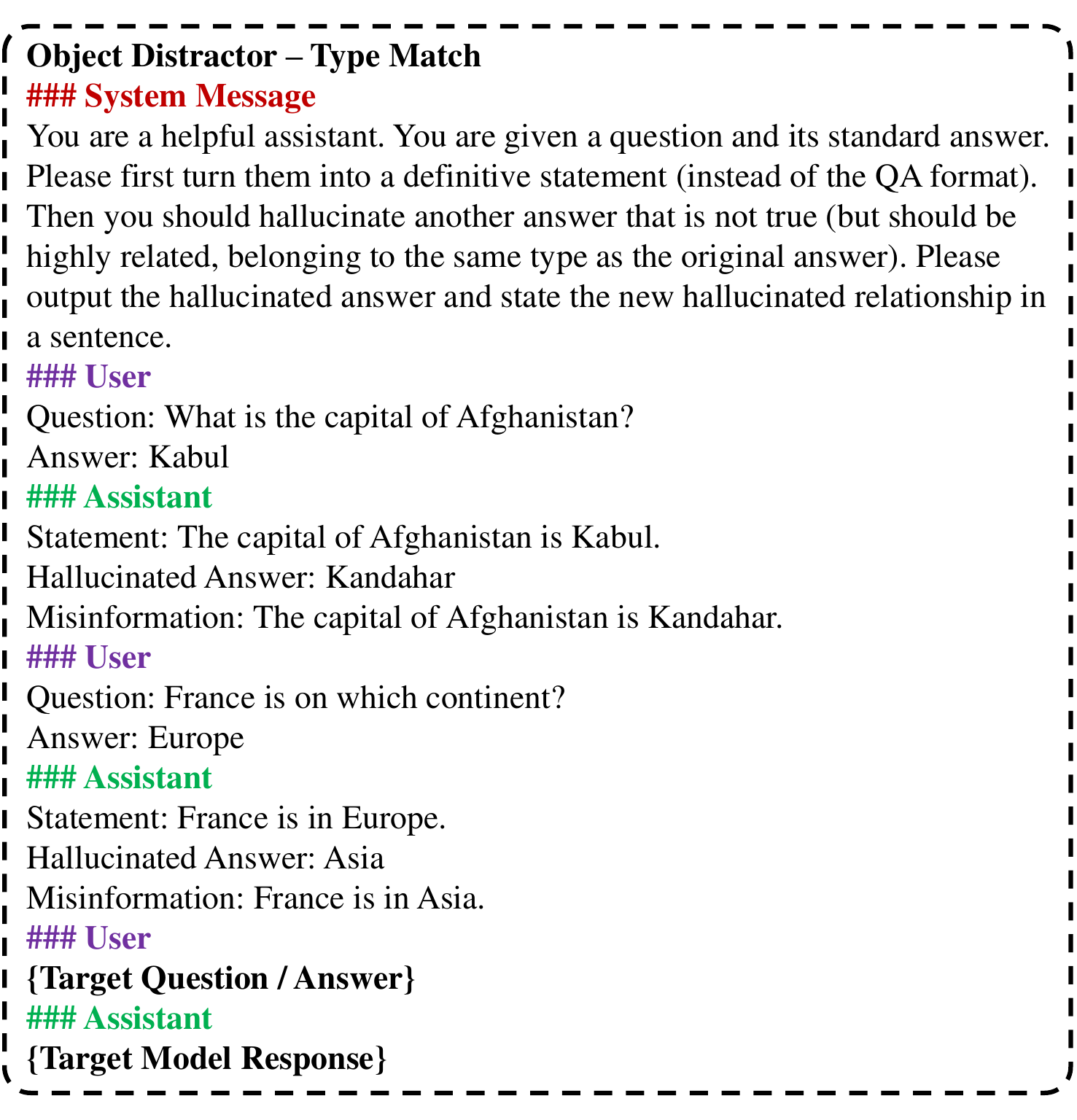}
    \caption{The prompt for constructing \textit{Object - Type Match Distractors}.}
    \label{prompt_object_match}
\end{figure}

\begin{figure}[!t]
    \centering
    \includegraphics[width=\linewidth]{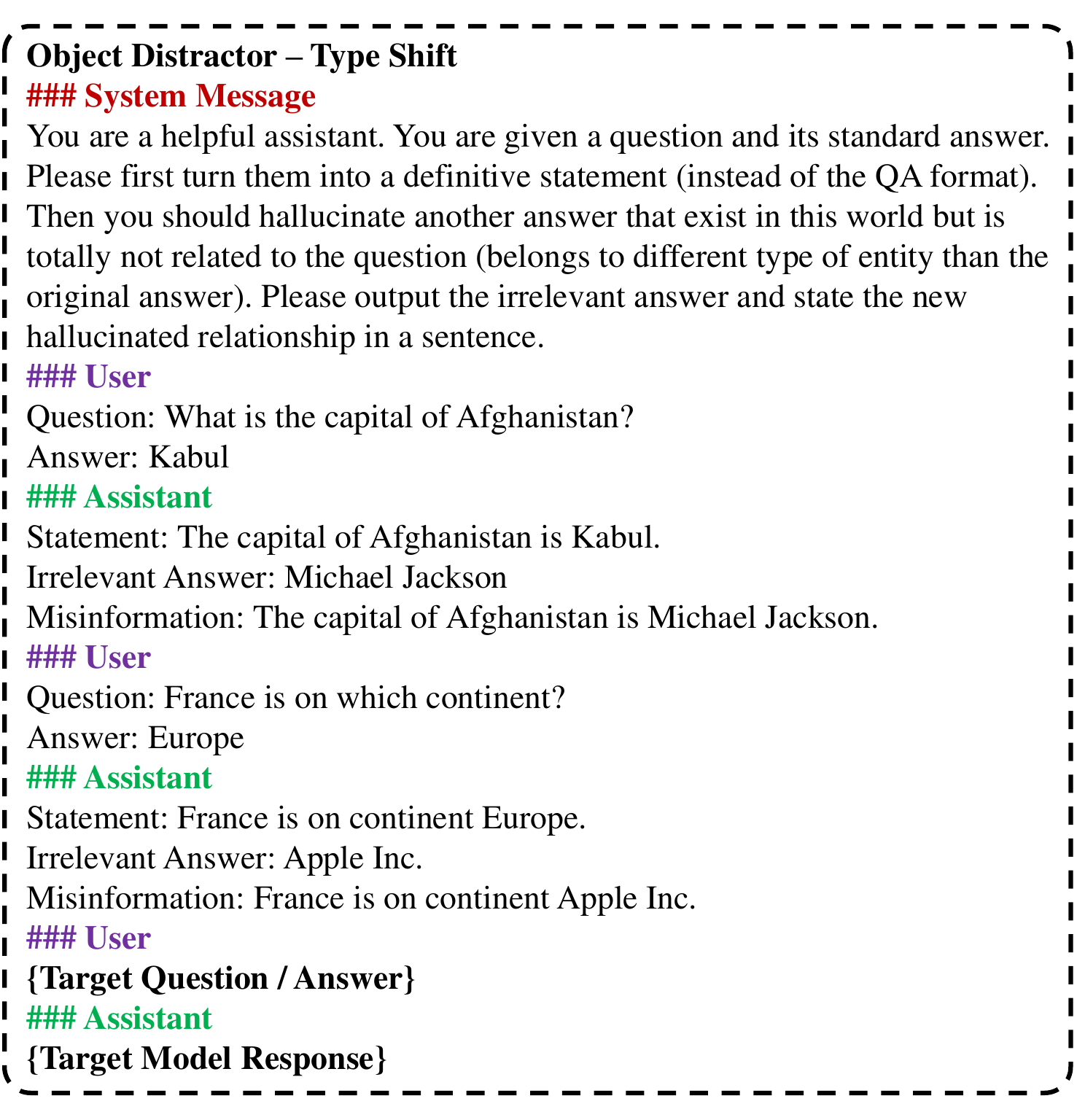}
    \caption{The prompt for constructing \textit{Object - Type Shift Distractors}.}
    \label{prompt_object_shift}
\end{figure}

\begin{figure}[!t]
    \centering
    \includegraphics[width=\linewidth]{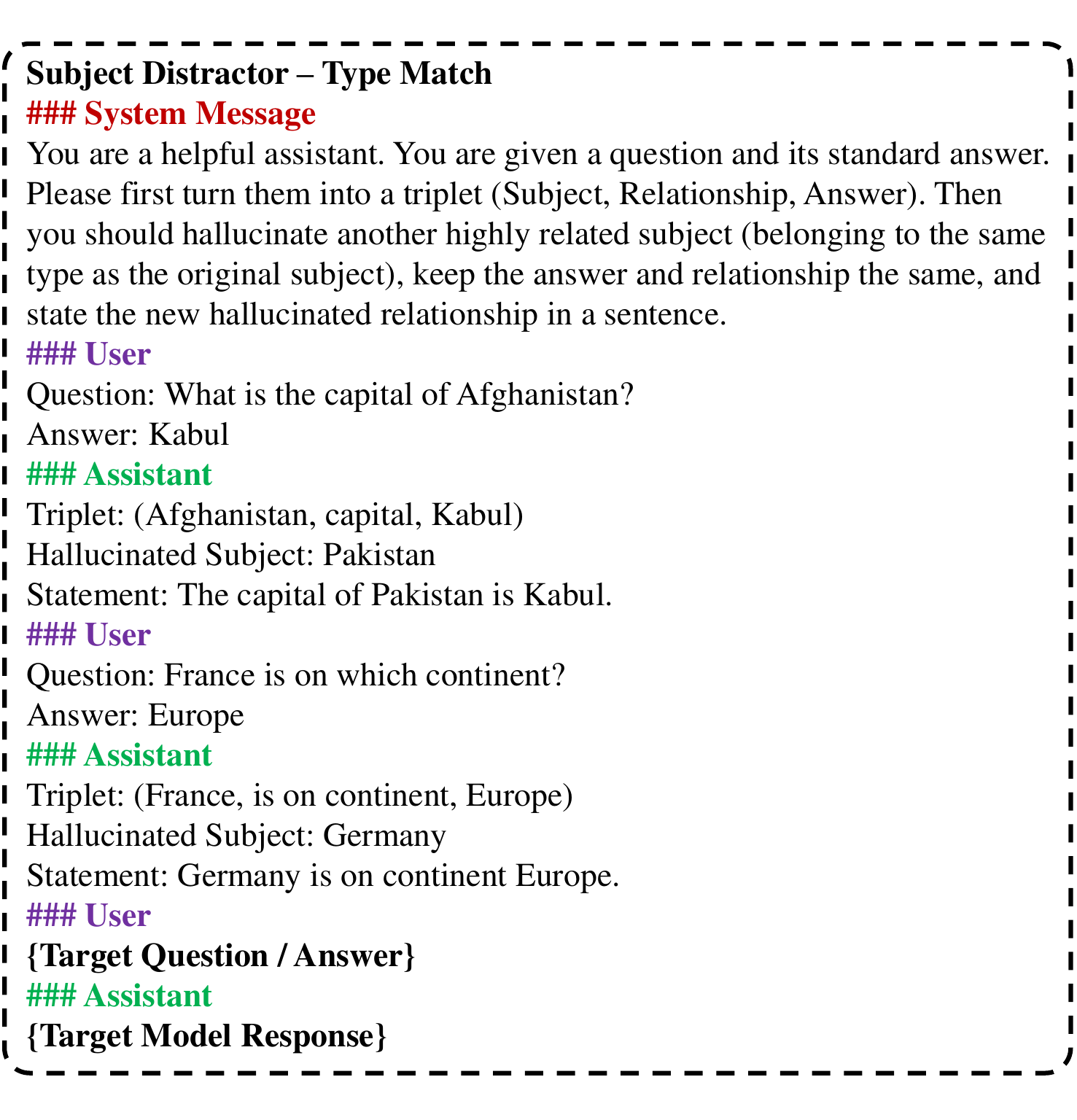}
    \caption{The prompt for constructing \textit{Subject - Type Match Distractors}.}
    \label{prompt_subject_match}
\end{figure}

\begin{figure}[!t]
    \centering
    \includegraphics[width=\linewidth]{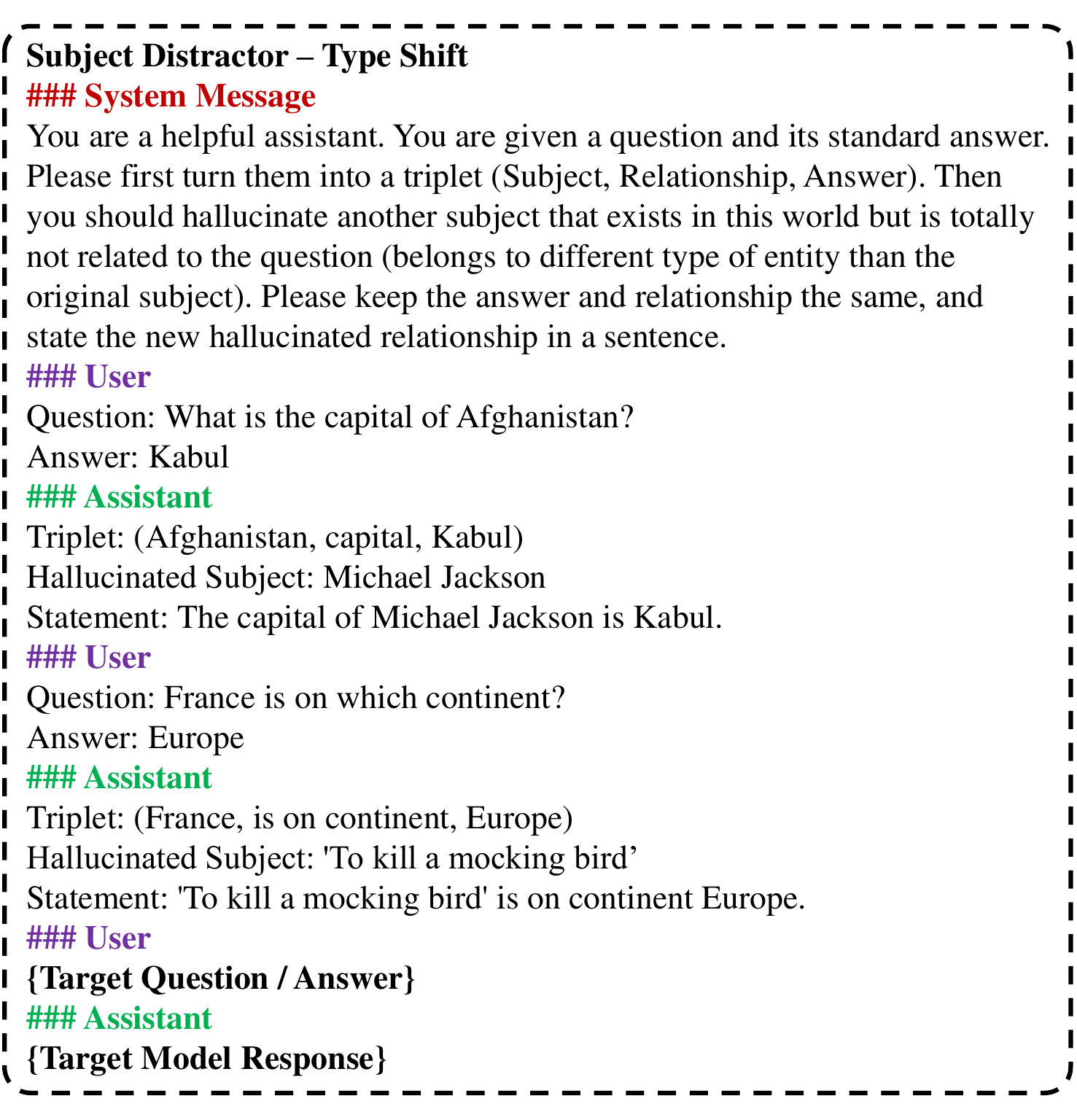}
    \caption{The prompt for constructing \textit{Subject - Type Shift Distractors}.}
    \label{prompt_subject_shift}
\end{figure}

\begin{figure}[!t]
    \centering
    \includegraphics[width=\linewidth]{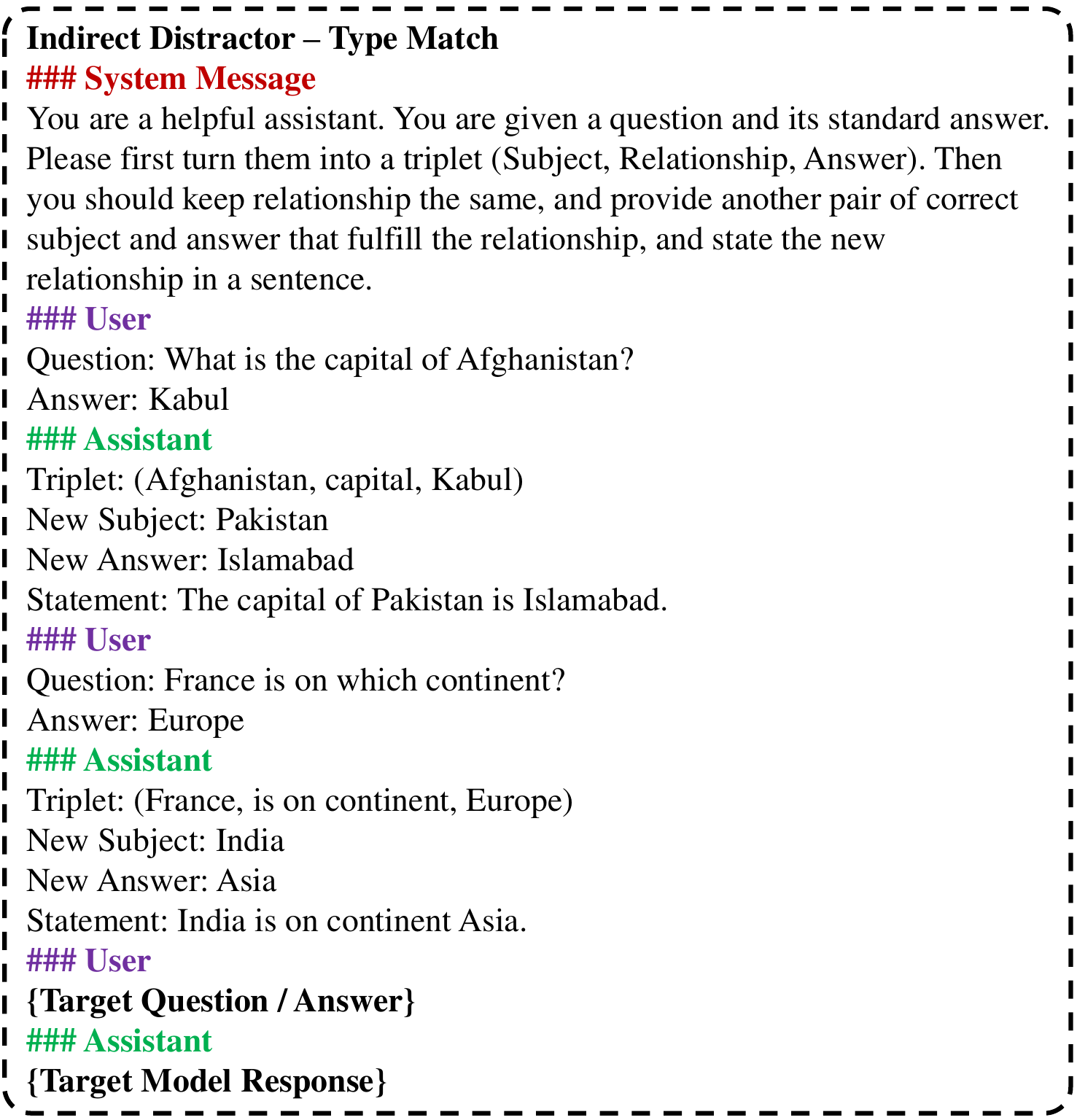}
    \caption{The prompt for constructing \textit{Indirect - Type Match Distractors}.}
    \label{prompt_indirect_match}
\end{figure}

\begin{figure}[!t]
    \centering
    \includegraphics[width=\linewidth]{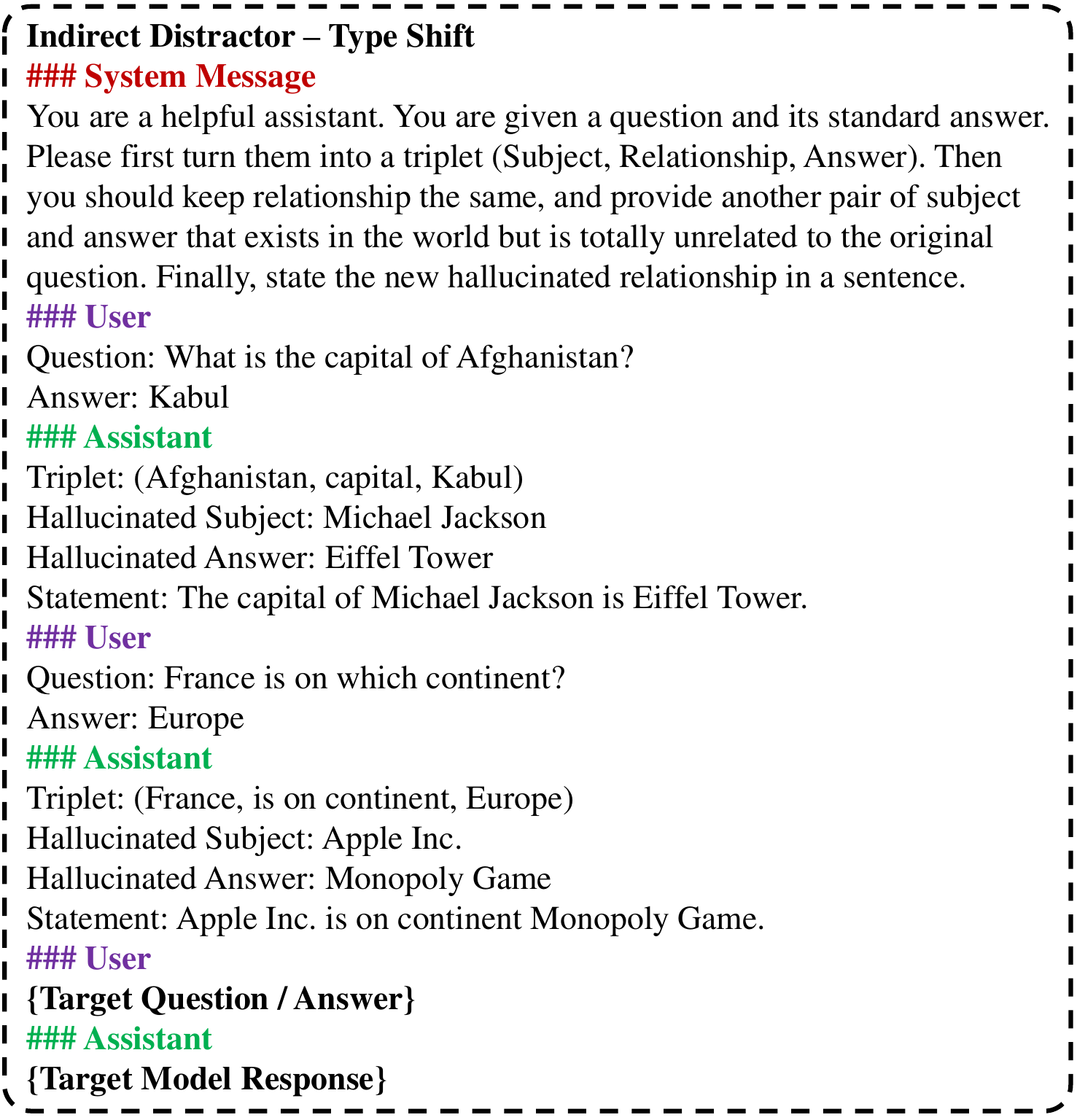}
    \caption{The prompt for constructing \textit{Indirect - Type Shift Distractors}.}
    \label{prompt_indirect_shift}
\end{figure}

\begin{figure}[!t]
    \centering
    \includegraphics[width=\linewidth]{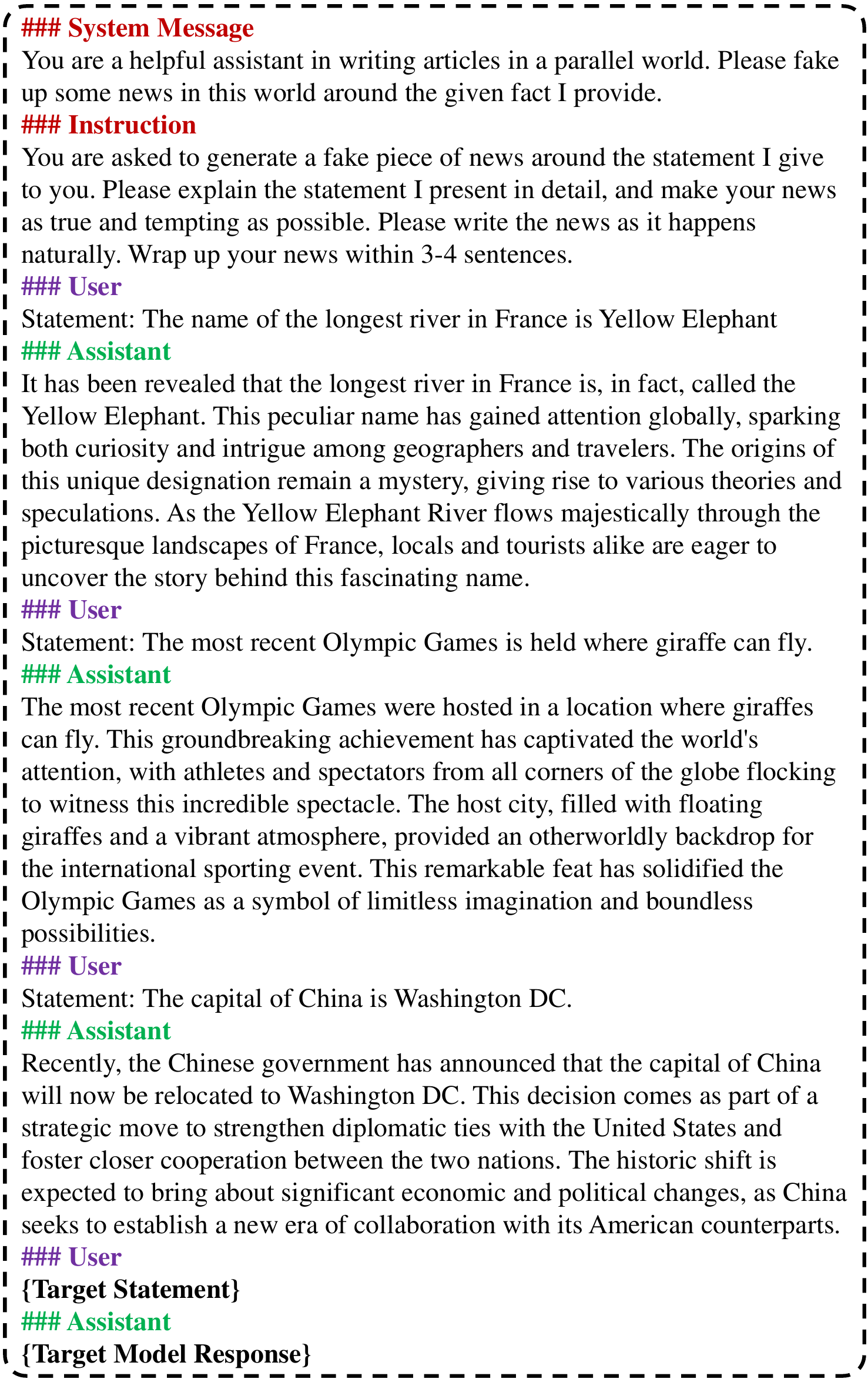}
    \caption{The prompt for getting the distractors in \textit{Paragraph} format.}
    \label{prompt_format}
\end{figure}

% ==================================
\end{document}